\documentclass[11pt]{article}

\usepackage[margin=1in]{geometry}

\usepackage{xcolor}
\definecolor{navy}{rgb}{0, 0, 0.75}

\usepackage{graphicx}
\usepackage{tikz}

\usepackage{subcaption}

\usepackage{amsmath, amsthm, amssymb}
\usepackage{bm}

\usepackage{mathrsfs}

\usepackage[shortlabels]{enumitem}

\usepackage{keytheorems}

\usepackage{todonotes}
\usepackage[colorlinks=true,linktoc=page,allcolors=navy]{hyperref}
\usepackage[capitalize,nosort]{cleveref}

\AddToHook{cmd/appendix/before}{\crefalias{section}{appendix}}

\theoremstyle{plain}
\newkeytheorem{theorem}[
  parent = section,
  refname = {Theorem,Theorems},
  Refname = {Theorem,Theorems}
]
\numberwithin{theorem}{section}
\newkeytheorem{corollary}[
  sibling = theorem,
  refname = {Corollary,Corollaries},
  Refname = {Corollary,Corollaries}
]
\newkeytheorem{lemma}[
  sibling = theorem,
  refname = {Lemma,Lemmas},
  Refname = {Lemma,Lemmas}
]

\theoremstyle{definition}
\newkeytheorem{definition}[
  sibling = theorem,
  refname = {Definition,Definitions},
  Refname = {Definition,Definitions}
]
\newkeytheorem{remark}[
  sibling = theorem,
  refname = {Remark,Remarks},
  Refname = {Remark,Remarks}
]
\newkeytheorem{example}[
  sibling = theorem,
  refname = {Example,Examples},
  Refname = {Example,Examples}
]
\newcommand{\E}{\mathbb{E}}
\renewcommand{\Pr}{\mathrm{Pr}}

\DeclareMathOperator*{\argmin}{argmin}
\DeclareMathOperator*{\argmax}{argmax}

\newcommand{\eR}{\overline{\R}}

\newcommand{\eps}{\varepsilon}
\renewcommand{\phi}{\varphi}

\newcommand{\rn}{\bm} 

\newcommand{\A}{\mathcal{A}}
\newcommand{\B}{\mathcal{B}}
\newcommand{\C}{\mathcal{C}}
\newcommand{\D}{\mathcal{D}}
\newcommand{\F}{\mathcal{F}}
\newcommand{\N}{\mathbb{N}}
\newcommand{\R}{\mathbb{R}}
\newcommand{\V}{\mathcal{V}}
\newcommand{\W}{\mathcal{W}}
\newcommand{\X}{\mathcal{X}}
\newcommand{\Y}{\mathcal{Y}}

\renewcommand{\le}{\leqslant}
\renewcommand{\ge}{\geqslant}

\newcommand{\abs}[1]{\lvert#1\rvert}
\newcommand{\bigabs}[1]{\bigl\lvert#1\bigr\rvert}
\newcommand{\Bigabs}[1]{\Bigl\lvert#1\Bigr\rvert}

\newcommand{\sign}{\mathrm{sign}}

\DeclareMathOperator{\dom}{\mathrm{dom}}
\DeclareMathOperator{\spn}{\mathrm{span}}

\usepackage{accents}

\newcommand{\ch}{W}
\newcommand{\CSM}{{\Delta_k^k}}
\newcommand{\kCSM}[1]{{\Delta_{#1}^{#1}}}

\newcommand{\x}{{\rn{x}}}
\newcommand{\y}{{\rn{y}}}

\newcommand{\eat}[1]{}

\usepackage{cancel}
\usepackage{nicefrac}

\usepackage{comment}

\title{Universal Hardness From Masking Channels}

\author{
    Parikshit Gopalan \\
    Apple
\and
    Manoj Prabhakaran\thanks{Work done partly while at Apple.} \\
    IIT Bombay
\and
    Pranay Tankala\thanks{Work done during an internship at Apple, as well as with support from Simons Foundation Grant 733782 and Cooperative Agreement CB20ADR0160001 with the United States Census Bureau.} \\
    Harvard
}

\date{November 2, 2026}

\begin{document}

\maketitle
\thispagestyle{empty}

\begin{abstract}
    We prove a broad learning-theoretic generalization of the celebrated \emph{hardcore lemma} concerning the average-case complexity of computing a Boolean function (Impagliazzo, 1995). Our starting point is the observation that the classic notion of a hardcore measure, which reweights a function's input distribution, is equivalent to an input-dependent \emph{masking channel} that injects randomness into the function's output distribution until it appears fully pseudorandom. The analogue of density of the measure is the {\em fidelity} of the channel; a high-fidelity channel injects very little noise. 
    From this viewpoint, we develop a general framework for certifying hardness that allows for non-Boolean functions $f$ and a wide range of notions of hardness besides the classic 0-1 loss, by choosing the right notion of fidelity via convex duality. We show the difficulty of learning $f$ using a class $\C$ to minimize a convex loss is tightly characterized by the maximum fidelity possible for a masking channel, which measures the least amount of noise the channel must add to hide the structure of $f$ and make it $\C$-indistinguishable from random noise.  
    
    We derive the following concrete applications from this general framework: 
    \begin{itemize}
    \item We instantiate our framework to derive a number of multiclass generalizations of the hardcore lemma, and reveal some unexpected subtleties in going to larger alphabets. We tightly characterize the hardness of learning with various losses in the binary setting. 
    
    \item We prove the existence of a \emph{universal} masking channel that applies to many natural notions of hardness and fidelity simultaneously. Our results show that such channels are dual to omnipredictors \cite{omnipredictors}, which optimize multiple losses at once. 

    \item We observe a symmetry between masking channels and generalizations of the hardcore lemma using pseudoentropy. We present a tight characterization of pseudoentropy using the MaxEnt algorithm from machine learning.   
    \end{itemize}

    A long line of research has shown that learning and hardness are two sides of the same coin; our framework gives a crisp and general formulation of this principle via convex duality.
\end{abstract}

\newpage
\thispagestyle{empty}
\tableofcontents
\newpage
\pagestyle{plain}
\setcounter{page}{1}

\section{Introduction}

Given a function class $\C$, which might comprise machine learning models of a given architecture or Boolean circuits of a given complexity, why might a Boolean function $f : \{\pm 1\}^n \to \{\pm 1\}$ be hard to approximate by functions in the class? In other words, what properties of $f$ might explain why every $c \in \C$ errs on at least a $\delta$-fraction of inputs to $f$?

One possible explanation is that the function $f$ has a \emph{hardcore} set $S \subseteq \{\pm 1\}^n$ of density at least $2\delta$ on which every circuit in $\C$ errs with probability close to $1/2$, the same error as achieved by naive random guessing. The classic \emph{hardcore lemma} in complexity theory states that this is, in fact, the only possible explanation. The lemma can be stated as follows, where we set $\X = \{\pm 1\}^n$ and $y = f(x)$:

\begin{theorem}[Hardcore Lemma \cite{impagliazzo1995hard,holenstein2005key}]
\label{thm:classic-hardcore}
    For all distributions over $\rn{x} \in \X$, functions $f : \X \to \{\pm 1\}$, classes $\C \subseteq \{\X \to \{\pm 1\}\}$, and $\delta, \eps \in (0, 1)$, at least one of the following holds:\footnote{Throughout this paper, we use boldface, as in ``$\rn{x}$'' and ``$\rn{y}$'', to indicate random variables.}
    \begin{itemize}
        \item \textbf{(Easy Case)} For some $c : \X \to \R$ that is the average of $O(\log(1/\delta)/\eps^2)$ functions in $\C$,
        \[
            \Pr[\sign(c(\rn{x}) - t) = f(\rn{x})] \ge 1 - \delta,
        \]
        where $t \in \R$ is some threshold. \item \textbf{(Hard Case)} There exists a measure $\mu : \X \to [0, 1]$ with the following properties:\footnote{In many cases, a $[0,1]$-valued \emph{hardcore measure} can be converted to a $\{0, 1\}$-valued indicator function, or equivalently, a \emph{hardcore set} \cite{impagliazzo1995hard}. In this paper, we do not belabor the distinction between these two objects.}
        \begin{itemize}
            \item \textbf{(Density)} $\E[\mu(\rn{x})] \ge 2\delta$,
            \item \textbf{(Hardcore)} $\E[\mu(\rn{x})\bm{1}[c(\rn{x}) = f(\rn{x})]] \le (1/2 + \eps)\E[\mu(\rn{x})]$ for all $c \in \C$.
        \end{itemize}
    \end{itemize}
\end{theorem}
This lemma was first proved by Impagliazzo in \cite{impagliazzo1995hard} with density $\delta$. Holenstein \cite{holenstein2005key} improved the density to the tight bound of $2\delta$.  

\paragraph{From Measures to Noisy Channels.}
In this paper, we propose a new generalization of the hardcore lemma approach to certifying hardness beyond the setting described in \cref{thm:classic-hardcore}, which will naturally extend to settings where $f$ is not necessarily Boolean-valued and there are a variety of loss functions of interest beyond 0-1 loss. To begin, let us consider a natural relaxation of the property that $f$ has a hardcore measure: that the output of $f$ can be made \emph{computationally indistinguishable} from pure random noise by passing it through a {\em masking} channel which flips the bit $f(x)$ with probability $\eta(x)$, where
\[
    \E[\eta(\rn{x})] \le \frac{1}{2} - \delta.
\]
Note that the noise rate $\eta(x)$ is allowed to depend on $x$, which allows an adversary to concentrate noise on the {\em easy parts} of the domain.
This relaxed property is directly implied by the hardcore property, and it directly implies that $f$ is $\delta$-hard against $\C$.\footnote{We ignore the small error term $\eps > 0$ in this informal discussion.}

\begin{figure}[t]
    \centering
    \includegraphics[width=0.45\linewidth]{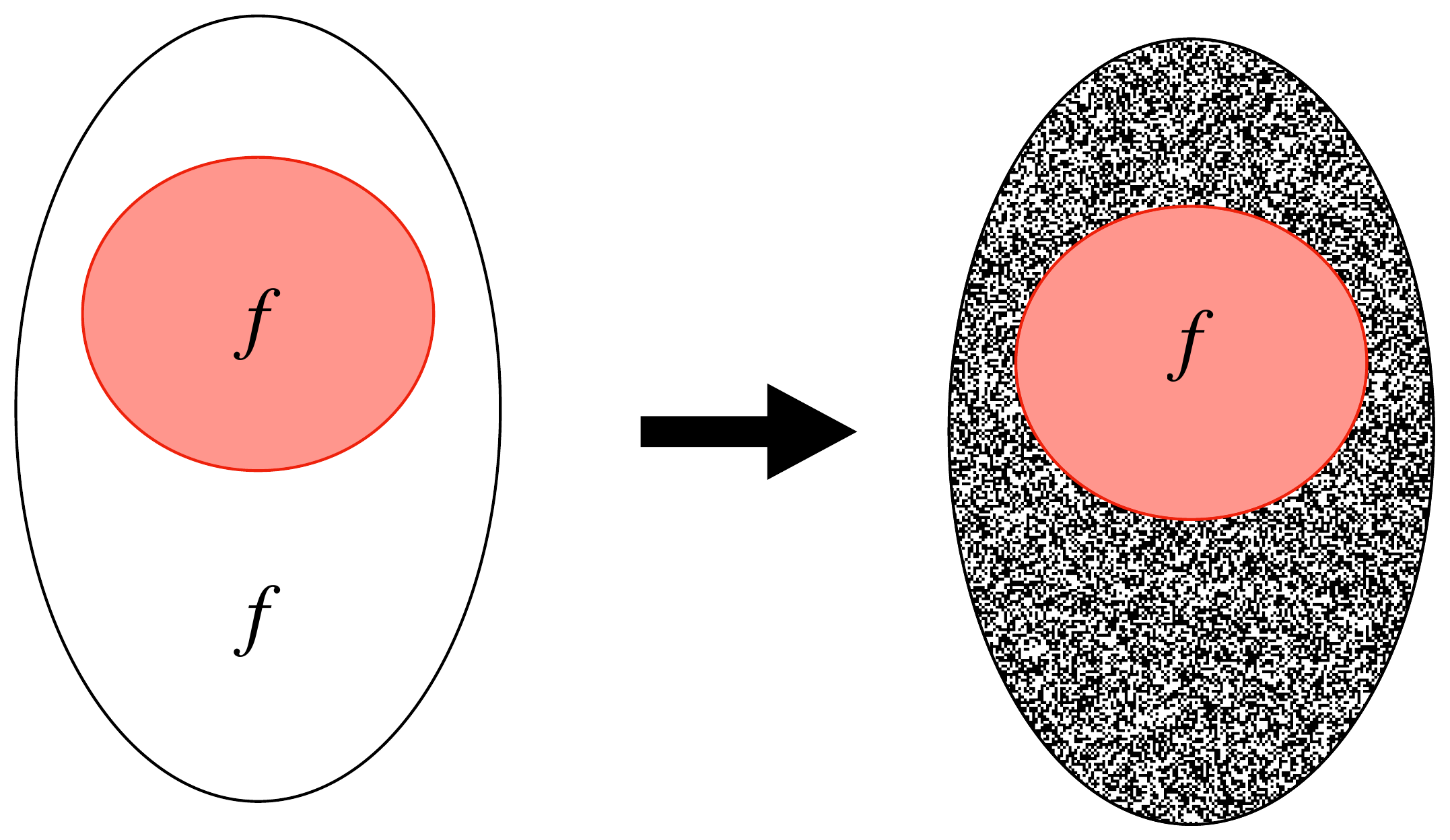}
        \caption{Masking $f$ by randomizing outside the hardcore set.}
        \label{fig:random_outside_S}
\end{figure}

Given a hardcore set $S \subseteq \X$, we can construct such a masking channel by replacing the function outside $S$ by random bits, as shown in Figure \ref{fig:random_outside_S}. For a distinguisher from $\C$ to tell the resulting labels apart from random noise, they would need to have an advantage over the set $S$ itself, which is not possible by the hardcore property. The larger the hardcore set, the less the noise needed in expectation. In contrast, for an easy function (say if $f \in \C$), the channel would need to randomize every point in the domain $\X$, or, more generally, corrupt half the labels on average. 
More generally, given a $2\delta$-dense hardcore measure $\mu$, it is easy to see that one such masking channel arises from the channel given by $\eta(x) = (1 - \mu(x))/2$, or equivalently from a noisy label $\tilde{\rn{y}}$ with 
$\E[\tilde{\rn{y}}|\rn{x}] = \mu(\rn{x})f(\rn{x})$.

Further, we claim that any such masking channel certifies the hardness of $f$.  The quantitative bound is better when the amount of noise added is small. To see this, recall that two functions $a, b : \X \to \{\pm 1\}$ are said to be \emph{$(\C, \eps)$-indistinguishable} \cite{TrevisanTV09} if
\(
    \max_{c \in \C}\,\abs{\E[c(\rn{x})(a(\rn{x}) - b(\rn{x}))} \le \eps,
\)
and that this definition extends easily to pairs of non-deterministic conditional distributions $\rn{a}|\rn{x}$ and $\rn{b}|\rn{x}$.\footnote{For simplicity, let us suppose that $\C$ is closed under negation, so that the absolute values become irrelevant.}
If we let $\rn{y},\tilde{\rn{y}},\rn{b} \in \{\pm 1\}$ denote $f(\rn{x})$, the noisy label, and an independent uniform bit, respectively, then for all $c \in \C$,
\[
    \Pr[c(\rn{x}) \neq \rn{y}] \ge \Pr[c(\rn{x}) \neq \tilde{\rn{y}}] - \E[\eta(\rn{x})] \approx_{\eps} \underbrace{\Pr[c(\rn{x}) \neq \rn{b}]}_{=1/2} - \underbrace{\E[\eta(\rn{x})]}_{\le 1/2 - \delta} \ge \delta.
\]

Our framework reinterprets the role of the measure $\mu$. While $\mu$ was originally conceived as a reweighting of the distribution over features $x \in \X$, it recasts $\mu$ as a noisy channel that leaves the marginal distribution over features fixed and instead modifies the conditional distribution over labels. This change in perspective was crucially used in distribution-specific agnostic boosting algorithms \cite{feldman2010boosting,kanade2009boosting}. We will see that it is extremely fruitful for certifying hardness.

\paragraph{Our Contributions.} In this work, we design a general framework, based on masking channels, for certifying the hardness of a function, under various notions of what it means for functions to be easy, or have structure that is identifiable by $\C$, that substantially generalizes and extends the hardcore lemma for Boolean functions and 0-1 loss.  From a machine learning point of view, this generalization gives tight, quantitative characterizations of the best loss one might hope to achieve when fitting a distribution over labeled data points $(\rn{x}, \rn{y})$ using models from a prescribed class. From a computational complexity point of view, we give characterizations of the average-case complexity of a function, with respect to a wide range of notions of hardness for functions $f : \X \to \Y$, where $\abs{\Y} = k$ is finite. In particular:

\begin{enumerate}[(1)]
    \item \textbf{(\cref{sec:intro-hardness-fidelity})}
    Our main conceptual contribution is a shift in focus from hardcore measures $\mu : \X \to [0, 1]$ to hardness certificates based on input-dependent \emph{masking channels}
    \[
        \ch : \X \to \R^{k \times k}.
    \]
    Here, $\ch(x)$ is a column-stochastic matrix mapping each possible label $y \in \Y$ to a distribution $\ch(x)y \in \Delta_k$ over noisy labels $\tilde{y} \in \Y$, all represented as $k$-dimensional column vectors. Suppose the noisy label distribution $\ch(\rn{x})\rn{y} \in \Delta_k$ is $\C$-indistinguishable from the uniform distribution $\vec{1}/k$ when labeled data points $(\rn{x}, \rn{y})$ are drawn from a given distribution $\D$. We call such a channel a {\em masking channel} for $\C$ under the distribution $\D$ and say that the output of the masking channel is $\C$-pseudorandom.
    
    \item \textbf{(\cref{sec:intro-hardness-fidelity} and \cref{sec:channels})} We show that the average-case \emph{hardness} of approximating the labels using $\C$ is equal to the highest \emph{fidelity} masking channel $\ch$ with $\C$-pseudorandom outputs, provided that the chosen notions of ``hardness'' and ``fidelity'' are \emph{convex-concave conjugates}. In other words, hardness is related to the \emph{least} amount of noise that must be added to the function's output to mask the signal and make it $\C$-pseudorandom. Our theorem, which crucially relates pseudorandomness constraints to $\ell_1$ regularization via convex duality, can be viewed as sharpening a complexity-theoretic result via a learning-theoretic perspective.
    
    \item \textbf{(\cref{sec:intro-applications} and \cref{sec:applications})} We show that our hardness-fidelity theorem implies the existence of several distinct and interesting hardness certificates in the multiclass setting. In particular, by combining our framework with new insights on surrogate losses for the 0-1 loss, we uncover a landscape of novel ``hardcore lemmas'' in the multiclass setting.
    
    \item \textbf{(\cref{sec:intro-universal} and \cref{sec:universal})} Using multicalibration and related ideas, we construct a \emph{universal masking channel}, which serves as a nearly tight certificate of the hardness of a function $f$ with respect to a wide range of different notions of hardness \emph{simultaneously}. A priori, the existence of such a universal channel is very surprising.

    \item \textbf{(\cref{sec:intro-maxent} and \cref{sec:maxent})} Finally, we observe an elegant symmetry between our results and existing results on \emph{generalized pseudoentropy} in complexity theory. We show that \emph{MaxEnt algorithm} in machine learning gives a learning-theoretic exact characterization of pseudoentropy.
\end{enumerate}

\subsection{Hardness Equals Best Masking Channel Fidelity}
\label{sec:intro-hardness-fidelity}

Before setting up our framework of masking channels and channel fidelity, we give a simple example of what the framework has to offer. To begin, let us consider the optimal $(\C,\eps)$-hardcore density $d^\star$ in the binary setting which can be expressed with the following optimization problem:\footnote{The constraint $\abs{\E[\mu(\rn{x})c(\rn{x})\rn{y}]} \le \eps$ differs slightly from the definition of a hardcore measure that we gave in \cref{thm:classic-hardcore}, which can be expressed as $\E[\mu(\rn{x})c(\rn{x})\rn{y}] \le 2\eps \E[\mu(\rn{x})]$. Ultimately, our general theorem will be flexible enough to handle both types of constraints and more.}
\begin{equation}
\label{eq:intro-hardcore}
    d^\star = \max_{\mu : \X \times \{\pm 1\} \to [0, 1]}\, \E[\mu(\rn{x}, \rn{y})]  \quad \text{s.t.} \quad \max_{c \in \C}\, \bigabs{\E[\mu(\rn{x}, \rn{y})c(\rn{x})\rn{y}]} \le \eps.
\end{equation}
The optimization problem above is well-defined for any fixed distribution over $(\rn{x}, \rn{y}) \in \X \times \{\pm 1\}$. In the special case that $\rn{y} = f(\rn{x})$ for a deterministic function $f$, we may assume without loss of generality that $\mu(x, y)$ ignores its second input and is a function $\mu(x)$ of $x$ alone.

Our framework will connect $d^\star$ to a different optimization problem, which at first seems entirely unrelated. This alternate problem attempts to minimize $\ell_1$-regularized \emph{hinge loss} attainable by the slightly enlarged class $\spn(\C)$ under the same distribution over $(\rn{x}, \rn{y})$ pairs. Let us use the abbreviation $c_\lambda(x) = \sum_i \lambda_i c_i(x)$ for the linear combination of functions in $\C \subseteq \{\X \to \R\}$ with coefficient vector $\lambda$. Define the hinge loss for a label $y \in \{\pm 1\}$ and action $a \in \R$ by the standard formula $\ell_{\mathrm{hinge}}(y, a) = (1 - ay)_+$, where $t_+ = \max(t, 0)$. Then, the value we seek is:
\begin{equation}
\label{eq:intro-hinge}
    p^\star = \min_{\lambda \in \R^\C}\, \E[(1 - c_\lambda(\rn{x})\rn{y})_+] + \eps \|\lambda\|_1.
\end{equation}

Soon, we will see that $p^\star = d^\star$, and we claim that this equality immediately implies the optimal $2\delta$-density version of the classic binary hardcore lemma. Indeed, it is well known in machine learning that the hinge loss of an action $a \in \R$ for a label $y \in \{\pm 1\}$ exceeds its 0-1 loss by at least a factor of $2$ after appropriate rounding (\cite{KKMS08}, also see \cref{sec:intro-applications} and \cref{rmk:01-vs-hinge}). Therefore, the inequality $p^\star \le d^\star$ yields the following chain of implications:
\begin{gather*}
    \text{$f$ is $\delta$-hard under $\ell_{\text{0-1}}$ for $\mathrm{thr}(\spn(\C))$} \\
    \Downarrow \\
    \text{$f$ is $2\delta$-hard under $\ell_{\text{hinge}}$ for $\spn(\C)$} \\
    \Downarrow \\
    \text{$f$ has a $2\delta$-dense hardcore with respect to $\C$.}
\end{gather*}
Remarkably, we see that the elusive factor of $2$, which was absent in the original version of the theorem \cite{impagliazzo1995hard} and only appeared much later \cite{holenstein2005key}, arises \emph{precisely} because of the factor of $2$ relating 0-1 loss to hinge loss.\footnote{Complexity-theoretic (as opposed to learning theoretic) statements of the hardcore lemma, such as \cref{thm:classic-hardcore}, typically take $\C$ to be the set of Boolean circuits of size at most $s$, and they assume that $f$ is $\delta$-hard under $\ell_{\text{0-1}}$ for Boolean circuits of size at most $rs$, for some scaling factor $r > 1$ (see \cref{thm:classic-hardcore}). Technically, functions in $\mathrm{thr}(\spn(\C))$ need not be computable by circuits of this size unless we verify that there exists a \emph{sparse} vector $\lambda$, with roughly $r$ or fewer nonzero coefficients, that is $\eps$-approximately optimal for the optimization problem in \eqref{eq:intro-hinge}. This is indeed true, with $r = \mathrm{poly}(1/\eps)$, but we omit the proof since it is not the focus of this work.} The factor-2 bound on the gap between the hinge loss and 0-1 loss was shown in the work of \cite{KKMS08} on agnostic learning, there it was used to give an agnostic learning algorithm with optimal error, rather than twice the optimal error.

In this paper, as we explore multiclass analogues of the hardcore lemma, this \emph{surrogate loss} point of view will turn out to be quite fruitful.

\paragraph{Masking Channels.}

In the multiclass case, the label $y$ is an element of a size-$k$ set $\Y$, which we view interchangeably as either $[k] = \{1, \ldots, k\} \subseteq \N$ or the standard basis vectors $e_1, \ldots, e_k \in \R^k$ (a.k.a. ``one-hot encodings''). We are given a distribution $\D$ over labeled data points $(\rn{x}, \rn{y}) \in \X \times \Y$. Our goal is to either find a structured explanation for how the labels $\rn{y}$ are generated from $\rn{x}$, or certify that $\rn{y}|\rn{x}$ is pseudorandom.  In some cases, we view $y = f(x)$ as a deterministic function of the input $x$, and in other cases, we consider an arbitrary joint distribution over input-output pairs, which we shall often refer to as \emph{feature-label pairs}.

Instead of measures $\mu : \X \to [0, 1]$, we now study input-dependent noisy channels $\ch$, which map a given label $y$ to an $x$-dependent distribution $\ch(x)y \in \Delta_k$ over noisy labels.

\begin{definition}[Noisy Channel \cite{coverT}]
    We let $\Delta_k \subseteq \R^k$ denote the \emph{probability simplex} of vectors with nonnegative coordinates summing to $1$. We similarly let $\CSM \subseteq \R^{k \times k}$ denote the set of $k$-by-$k$ \emph{column-stochastic matrices} with columns in $\Delta_k$.  Given finite sets $\X$ and $\Y = [k]$, a \emph{noisy channel} is simply a function
    \(
        \ch : \X \to \CSM.
    \)
\end{definition}

Given a deterministic function $f:\X \to \Y$, a noisy channel $\ch$ can be used to define a corrupted version of $f$, where $y=f(x)$ (one-hot encoded) is replaced with $\ch(x)y$ (a probability distribution).
To study the hardness of a function $f$ w.r.t.\ a function class $\C$, we now introduce the notion of a \emph{masking channel} for $f$, which corrupts $f$ into a randomized function that appears totally random to $\C$. Below we define a masking channel not for a deterministic function $f$, but for a distribution $\D$ over $\X\times\Y$ (allowing for randomized functions, and incorporating a distribution on the inputs).

\begin{definition}[Masking Channel]
\label{def:hardcore-channel}
   
    Given a distribution $\D$ over labeled data points $(\rn{x}, \rn{y}) \in \X \times \Y$, a finite collection of functions $\C \subseteq \{\X \to \R^k\}$, a noisy channel $\ch : \X \to \CSM$ and $\eps \ge 0$, we say that $W$ is a 
    \emph{$(\C, \eps)$-masking channel} for the distribution $\D$, if for all $c \in \C$,
    \[
        \bigabs{\E[\langle c(\rn{x}), \, W(\rn{x})\rn{y} - \vec{1}/k \rangle]} \le \eps.
    \]
\end{definition}

We say that a random variable $\rn{z}$ that is distributed jointly with $\rn{x}$ is $(\C, \eps)$-pseudorandom if for every $c \in \C$, we have $
        \bigabs{\E[\langle c(\rn{x}), \, \rn{z} - \vec{1}/k \rangle} \le \eps$.  
Thus the definition of a masking channel for $\D$ requires 
the induced label distribution $\ch(\rn{x})\rn{y}$ is \emph{$(\C, \eps)$-pseudorandom}. If the distribution $\D$ is clear from the context, we will suppress $\D$ and refer to $W$ as a $(\C, \eps)$-masking channel.

\begin{example}
\label{ex:mask}
For a simple example of such a masking channel, suppose that the label $\rn{y}$ is independent of the features $\rn{x}$ and has the following distribution:
\[
    \E[\rn{y} \mid \rn{x}] = \begin{pmatrix}
        1/2 \\
        1/2 \\
        0 \\
        \vdots \\
        0
    \end{pmatrix} \in \Delta_k.
\]
Then, one possibility for the channel $\ch$ is to redistribute some of the mass from the labels $1$ and $2$ to each of the remaining $k-2$ labels, until the marginal distribution is uniform:
\[
    \ch(x) = \begin{pmatrix}
        2/k & 0 & 0 & 0 & \cdots & 0 \\
        0 & 2/k & 0 & 0 & \cdots & 0 \\
        1/k & 1/k & 1 & 0 & \cdots & 0 \\
        1/k &1/k & 0 & 1 & \cdots & 0 \\
        \vdots & \vdots & \vdots & \vdots & \ddots & \vdots \\
        1/k & 1/k & 0  & 0 & \cdots & 1
    \end{pmatrix} \in \CSM.
\]
Since $\E[\ch(\rn{x})\rn{y} \mid \rn{x}] = \vec{1}/k$, the pseudorandomness condition is clearly satisfied for any $\C$ and $\eps \ge 0$. As this example illustrates, if a label $i \in [k]$ occurs with probability $0$, then the corresponding column of $W$ can be defined arbitrarily; in particular, if $y = f(x)$ for a deterministic function $f : \X \to [k]$, then there is only a single relevant column of $W(x)$ for each $x \in \X$.
\end{example}

\paragraph{Channel Fidelity.} In the binary setting, we sought the hardcore measure $\mu$ with the highest possible density. By analogy, in the multiclass setting, we shall seek the highest possible \emph{fidelity} among masking channels. Intuitively, a channel has high fidelity if it adds little noise to the original label. In various scenarios, there are many reasonable ways to quantify a channel's fidelity, and we would like a definition that can capture them all. In the following definition, we use the abbreviation $\eR = \R \cup \{\pm \infty\}$ for the set of extended real numbers.

\begin{definition}[Fidelity Function]
    Given a finite label set $\Y = [k]$, a \emph{fidelity function} is any function $h : \Y \times \Delta_k \to \eR$ that is closed, proper, and concave in its second input (\cref{def:convex}).
\end{definition}

\begin{example}
\label{ex:fid-flip}
    The simplest example of a fidelity function, which we will use repeatedly, is
    \[
        h(y, z) = \langle y, z \rangle = 1 - d_{\mathrm{TV}}(y, z).
    \]
    This function takes as input the original label $y = e_i$ and a noisy label distribution $z \in \Delta_k$, and outputs the scalar $z_i$, which is the probability that the label is not corrupted. Note that this fidelity function decreases as the total variation distance $d_{\mathrm{TV}}$ between $z$ and $y$ (a point mass) increases. Eventually, we will plug in $z = W(x)y$, where $W(x)$ is a column-stochastic channel matrix.
\end{example}

\begin{example}
\label{ex:fid-ce}
    Another example of a fidelity function is the negated cross-entropy or log-loss
    \[
        h(e_i, z) = \log z_i.
    \]
    Note that this fidelity function outputs $-\infty$ when $z_i = 0$, which is a valid output in $\eR$.
\end{example}

\begin{example}
\label{ex:fid-mix}
    For yet another example of a fidelity function, consider
    \[
        h(y, z) = \begin{cases}
            \mu &\text{if $z = \mu \cdot y + (1 - \mu) \cdot (\vec{1}/k)$ for some $\mu \in [0, 1]$}\\
            -\infty &\text{otherwise.}
        \end{cases}
    \]
    This fidelity function equals the one in the first example when the noisy label distribution $z$ has a specific form: it mixes the original label $y$ with the uniform distribution, with some mixing parameter $\mu \in [0, 1]$.\footnote{As we will later see, it is not by accident that we have used the same variable $\mu$ here as for hardcore measures.} If the noisy label distribution does not have this form, then this fidelity function penalizes it by assigning a fidelity value of $-\infty$.
\end{example}

\paragraph{Fidelity of Masking Channel Certifies Hardness of Learning.} Our first main result states:

\begin{quote}
\em
    The highest possible fidelity of a masking channel for a distribution exactly characterizes how hard it is to learn the distribution.
\end{quote}

More precisely, given 
a fidelity function $h$, there is a convex conjugate loss function (see \cref{sec:prelim-convex}) $g:\Y\times \R^k \to \eR$, such that the highest fidelity of a $(\C,\eps)$-masking channel for $\D$ \emph{equals} the smallest loss incurred in representing $\D$ by a linear combination of functions from $\C$ (with an $\ell_1$-regularization penalty weighted by $\eps$).

In what follows, and in the rest of this paper, the abbreviation $\pi(a) = a - \langle a, \vec{1}/k\rangle \cdot \vec{1}$ denotes the ``centered'' mean-$0$ version of an action vector $a \in \R^k$. Also, recall from the discussion of equation~\eqref{eq:intro-hinge} that $c_\lambda$ is the function in $\spn(\C)$ with coefficients $\lambda$. We will prove the following theorem in \cref{sec:channels}.

\begin{theorem}[
    name = Hardness vs Fidelity,
    label = thm:channels,
    store = thm:channels
]
    For all finite $\X$, $\Y = [k]$, distributions $\D$ over $(\rn{x}, \rn{y}) \in \X \times \Y$, finite $\C \subseteq \{\X \to \R^k\}$, convex-concave conjugates $g : \Y \times \R^k \to \R$, $h : \Y \times \Delta_k \to \eR$, and $\eps \ge 0$,\footnote{When $\eps \neq 0$ and $g(\cdot, \pi(\cdot))$ is bounded below, the infimum on the left-hand side can be replaced with a minimum.}
    \[
        \inf_{\lambda \in \R^\C}\, \E[g(\rn{y}, \pi(c_\lambda(\rn{x})))] + \eps \|\lambda\|_1 =
        \max_{\ch : \X \to \CSM}\, \E[h(\rn{y}, \ch(\rn{x})\rn{y})] \quad\text{s.t.}\;\; \text{ $\ch$ is $(\C, \eps)$-masking}.
    \]
\end{theorem}

A unifying insight that runs through several works on the hardcore lemma is that learning and hardness are intimately related. Theorem \ref{thm:channels} gives a particularly general and crisp formulation of this insight through the lens of convex duality.

\subsection{Multiclass Hardcore Lemmas}
\label{sec:intro-applications}

We now consider concrete instantiations of \cref{thm:channels}. Doing so will lead to a rich landscape of distinct hardcore-like objects in the multiclass setting, each natural and interesting in its own right, that capture hardness with respect to \emph{different} multiclass loss functions. We then connect these various hardness certificates back to the classic (multiclass) 0-1 loss via its convex surrogate losses. All the results described in this section will be proved in \cref{sec:applications}.

\paragraph{Binary Setting.}
As a warm-up, we state several instantiations of \cref{thm:channels} in the binary setting, with specific choices of convex-concave conjugate pairs. \Cref{thm:binary-hardcores} below includes three versions, captured by equations~\eqref{eq:binary-hinge-hardcore}, \eqref{eq:binary-l1-hardcore}, and \eqref{eq:binary-lp-hardcore}.

\begin{theorem}[
    name = {Hinge, $\ell_1$, and $\ell_p$ Hardcore Lemmas},
    label = thm:binary-hardcores,
    store = thm:binary-hardcores
]
    Fix a finite set $\X$, a distribution over pairs $(\rn{x}, \rn{y}) \in \X \times \{\pm1\}$, a finite $\C \subseteq \{\X \to \R\}$, and $\eps > 0$. Then, using the abbreviation $\rn{\mu} = \mu(\rn{x}, \rn{y})$,
    \begin{align}
        \label{eq:binary-hinge-hardcore}
        \min_{\lambda \in \R^\C}\, \E[(1 - c_\lambda(\rn{x})\rn{y})_+] + \eps \|\lambda\|_1
        &=
        \max_{\mu : \X \times \Y \to [0, 1]}\, \E[\rn{\mu}] \quad \text{s.t.} \quad \forall c \in \C,\; \bigabs{\E[\rn{\mu}c(\rn{x})\rn{y}]} \le \eps, \\
        \label{eq:binary-l1-hardcore}
        \min_{\lambda \in \R^\C}\, \E[\abs{\rn{y} - c_\lambda(\rn{x})}] + \eps \|\lambda\|_1
        &=
        \max_{\mu : \X \times \Y \to [-1, +1]}\, \E[\rn{\mu}] \quad \text{s.t.} \quad \forall c \in \C,\; \bigabs{\E[\rn{\mu}c(\rn{x})\rn{y}]} \le \eps, \\
        \label{eq:binary-lp-hardcore}
        \min_{\lambda \in \R^\C}\, \E\Bigl[\frac{1}{p}\abs{\rn{y} - c_\lambda(\rn{x})}^p\Bigr] + \eps \|\lambda\|_1
        &=
        \max_{\mu : \X \times \Y \to \R}\, \E\Bigl[\rn{\mu} - \frac{1}{q}\abs{\rn{\mu}}^q\Bigr] \quad \text{s.t.} \quad \forall c \in \C,\; \bigabs{\E[\rn{\mu}c(\rn{x})\rn{y}]} \le \eps,
    \end{align}
    where $p,q > 1$ are any real numbers satisfying $1/p + 1/q = 1$.
\end{theorem}

\noindent We defer the proof to \cref{sec:applications}, but discuss their interpretation now.

\begin{itemize}
\item First, equation~\eqref{eq:binary-hinge-hardcore} is the relationship between hinge loss and hardcore density discussed in the previous section.
\item Next, equation~\eqref{eq:binary-l1-hardcore} relates the minimum $\ell_1$ loss to the maximum expected value of a $[-1, +1]$-valued \emph{signed} measure. $\ell_1$ loss minimization followed by randomized rounding is the algorithm used in \cite{KKMS08} to agnostically learn functions that are well approximated by linear combinations of $\C$. Its dual, which allows $\mu$ to take values in the interval $[-1, +1]$ is quite natural; for example, multiplying $y$ by a factor of $-1$ corresponds to deterministically flipping the label.
\item Finally, equation~\eqref{eq:binary-lp-hardcore} characterizes hardness with respect to all $\ell_p$ losses for $p > 1$. In this case, $\mu$ is both a \emph{signed} and \emph{unbounded} measure. Although it is no longer forced to output values in $[-1, +1]$, it is nevertheless strongly encouraged to do so by a penalty term of $-(1/q)\abs{\rn{\mu}}^q$ in the objective for $q=p/(p-1)$, which is negligible when $\abs{\rn{\mu}} \ll 1$ and very large when $\abs{\rn{\mu}} \gg 1$.
\end{itemize}

\paragraph{Multiclass Setting.} Let us now discuss three natural ways to certify the hardness of a conditional label distribution $\rn{y}|\rn{x}$ in the $k$-class setting, all of which generalize the familiar notion of a hardcore measure in the binary case ($k = 2$). Using our framework, we show that each of these three distinct hardness certificates (i.e. ``hardcore-like'' objects) tightly characterizes hardness with respect to a distinct multiclass loss function.

First, one might seek the densest measure $\mu$ on which the labels are $\C$-indistinguishable from the uniform distribution $\vec{1}/k$, or equivalently, such that the noisy channel outputting the mixture $\mu(\rn{x},\rn{y}) \cdot \rn{y} + (1 - \mu(\rn{x},\rn{y})) \cdot \vec{1}/k$ is in fact a $\C$-masking channel. \Cref{thm:weak-hinge-hardcore} below shows that this density captures hardness with respect to what we call the \emph{weak hinge loss}, defined for $i \in [k]$ and mean-$0$ vectors $a \in \R^k$ by the formula
\(
    \ell_{\mathrm{weak}}(e_i, a) = (1 - a_i)_+.
\)
(If $a$ is not mean-$0$, we center it first.) We call this loss ``weak'' because it is relatively easy to minimize: if even a single label $i \in [k]$ can be eliminated with certainty, then $a = \vec{1} - ke_i$ achieves $0$ loss.
\begin{theorem}[
    name = {Weak Hinge Loss vs Uniform Noise},
    label = thm:weak-hinge-hardcore,
    store = thm:weak-hinge-hardcore
]
    For all finite $\X$ and $\Y = [k]$, distributions over $(\rn{x}, \rn{y}) \in \X \times \Y$, finite $\C \subseteq \{\X \to \R^k\}$, and $\eps > 0$,
    \[
        \min_{\lambda \in \R^\C}\, \E[\ell_{\mathrm{weak}}(\rn{y}, c_\lambda(\rn{x}))] + \eps \|\lambda\|_1 =
        \begin{aligned}[t]
            \max_{\mu : \X \times \Y \to [0, 1]}\, &\E[\mu(\rn{x}, \rn{y})] \quad \\\text{s.t.} &\; \forall c \in \C,\; \bigabs{\E[\mu(\rn{x},\rn{y})\langle c(\rn{x}), \, \rn{y} - \vec{1}/k\rangle]} \le \eps.
        \end{aligned}
    \]
\end{theorem}

Second, one might seek the least amount of (not necessarily uniform!) noise that must be added to the labels to achieve global indistinguishability from uniform over $[k]$. \Cref{thm:strong-hinge-hardcore} below shows that among such masking channels $\ch$, the maximum possible fidelity $\E[\langle \rn{y}, \ch(\rn{x})\rn{y}\rangle]$, which can also be written $1 - \E[d_{\mathrm{TV}}(\rn{y}, \ch(\rn{x})\rn{y})]$, captures hardness with respect to what we call the \emph{strong hinge loss}, defined for $i \in [k]$ and mean-$0$ vectors $a \in \R^k$ by the formula $\ell_{\mathrm{strong}}(e_i, a) = \max_{j \in [k]}(e_i - a)_j$. (If $a$ is not mean-$0$, center it first.) We call this loss ``strong'' simply because it is a pointwise upper bound on $\ell_{\mathrm{weak}}$.

\begin{theorem}[
    name = {Strong Hinge Loss vs Arbitrary Noise},
    label = thm:strong-hinge-hardcore,
    store = thm:strong-hinge-hardcore
]
    For all finite $\X$ and $\Y = [k]$, distributions over $(\rn{x}, \rn{y}) \in \X \times \Y$, finite $\C \subseteq \{\X \to \R^k\}$, and $\eps > 0$,
    \[
        \min_{\lambda \in \R^\C}\, \E[\ell_{\mathrm{strong}}(\rn{y}, \pi(c_\lambda(\rn{x})))] + \eps \|\lambda\|_1 =
        \begin{aligned}[t]
            \max_{\ch : \X \to \CSM}\, &\E[\langle \rn{y}, \ch(\rn{x})\rn{y}\rangle] \\\text{s.t.} &\; \forall c \in \C,\; \bigabs{\E[\langle c(\rn{x}), \, \ch(\rn{x})\rn{y} - \vec{1}/k\rangle]} \le \eps.
        \end{aligned}
    \]
\end{theorem}

Third, one might seek modified labels that are as statistically far as possible from the original labels while remaining computationally indistinguishable from them.\footnote{Later, in \cref{sec:intro-maxent}, we discuss a closely related fourth possibility: finding the \emph{highest-entropy} labels that are indistinguishable from the original, which has been studied extensively in recent work.} \Cref{thm:cs-hinge-hardcore} below shows that the maximum attainable statistical distance captures hardness with respect to the well-established \emph{Crammer-Singer hinge loss} \cite{crammer2001multiclass}, defined for $i \in [k]$ and $a \in \R^k$ by the formula $\ell_{\mathrm{CS}}(e_i, a) = (1 - a_i + \max_{j \neq i} a_j)_+$, which depends on the margin by which the score assigned to the correct label $i$ exceeds the maximum score assigned to any wrong label $j \neq i$.\footnote{Unlike the previous several results, \cref{thm:cs-hinge-hardcore} will not follow from \cref{thm:channels} itself, but rather from a more general version of it, \cref{thm:six}, which we state and prove in \cref{sec:general}.}
We emphasize that this statistical distance is measured from \emph{point mass} distribution $\rn{y} \in \{e_1, \ldots, e_k\}$, which depends on the coupling between original and noisy labels, as opposed to being measured from $\E[\rn{y}|\rn{x}] \in \Delta_k$.
\begin{theorem}[
    name = {Crammer-Singer Hinge Loss vs Maximum TV Distance},
    label = thm:cs-hinge-hardcore,
    store = thm:cs-hinge-hardcore
]
    For all finite $\X$ and $\Y = [k]$, distributions over $(\rn{x}, \rn{y}) \in \X \times \Y$, finite $\C \subseteq \{\X \to \R^k\}$, and $\eps > 0$,
    \[
        \min_{\lambda \in \R^\C}\, \E[\ell_{\mathrm{CS}}(\rn{y}, c_\lambda(\rn{x}))] + \eps \|\lambda\|_1 =
        \begin{aligned}[t]
            \max_{p : \X \times \Y \to \Delta_k}\, &\E[d_{\mathrm{TV}}(\rn{y}, p(\rn{x}, \rn{y}))] \quad \\\text{s.t.} &\; \forall c \in \C,\; \bigabs{\E[\langle c(\rn{x}), \, \rn{y} - p(\rn{x}, \rn{y})\rangle]} \le \eps.
        \end{aligned}
    \]
\end{theorem}

\paragraph{Surrogate Losses.}

At this point, we have shown that $\delta$-hardness under $\ell_{\mathrm{weak}}$ implies a $\delta$-density hardcore measure, that $\delta$-hardness under $\ell_{\mathrm{strong}}$ implies a hardcore channel with total variation distance at most $1- \delta$, and that $\delta$-hardness for $\ell_{\mathrm{CS}}$ implies  an indistinguishable model at total variation distance at least $\delta$. But each of these seems to be a different hardness assumption. In analogy to the binary hardcore lemma, one could ask: What certificates of hardness can we construct in the multiclass setting assuming only $\delta$-hardness under the 0-1 loss $\ell_{\text{0-1}}(y, \tilde{y}) = \bm{1}[y \neq \tilde{y}]$? Although this loss is not convex (for starters, because its action space is discrete), the previous consequences can be derived assuming $\ell_{\text{0-1}}$-hardness if we could show that $\ell_{\mathrm{weak}}, \ell_{\mathrm{strong}}, \ell_{\mathrm{CS}}$ were \emph{surrogate losses} for $\ell_{\text{0-1}}$ in the following sense:

\begin{definition}[$\beta$-surrogate Loss]
\label{def:surrogate}
    Given a label set $\Y=[k]$ and action sets $\A,\B$, and $\beta \ge 0$, we say that a loss function $\ell_\A : \Y \times \A \to \eR$ is a \emph{$\beta$-surrogate} for a loss function $\ell_\B : \Y \times \B \to \eR$ if $\B$ is finite and there exists a \emph{rounding function} $R : \A \to \Delta(\B)$ such that for all $y \in \Y$ and $a \in \A$,
    \[
        \ell_\A(y, a) \ge \beta \cdot \E_{\rn{b} \sim R(a)}[\ell_\B(y, \rn{b})].
    \]
\end{definition}

We show that whether or not a loss function, such as $\ell_{\mathrm{weak}}, \ell_{\mathrm{strong}}, \ell_{\mathrm{CS}}$, is a surrogate for the 0-1 loss can be completely determined from its associated \emph{entropy} function:

\begin{definition}[
    name = Entropy,
    label = def:entropy
]
Given a label set $\Y$, an action set $\A$, and a loss function $\ell : \Y \times \A \to \eR$, the \emph{entropy} of $\ell$ is the concave function $h : \Delta(\Y) \to \eR$ given by the formula
\[
    h(p) = \inf_{a \in \A}\, \E_{\rn{y} \sim p}[\ell(\rn{y}, a)].
\]
\end{definition}

\begin{theorem}[
    name = Entropy Surrogates,
    label = thm:opt-round,
    store = thm:opt-round
]
    Given sets $\A,\B,\Y$ and $\beta\ge 0$ and finite losses $\ell_\A : \Y \times \A \to \R$ and $\ell_\B : \Y \times \B \to \R$, let $h_\A, h_\B: \Delta(\Y) \to \eR$ denote their entropy functions. If $\Y,\B$ are finite, then $\ell_\A$ is a $\beta$-surrogate loss for $\ell_\B$ if and only if $h_\A(p) \ge \beta h_\B(p)$ for all $p \in \Delta(\Y)$.
\end{theorem}
An additive term can easily be incorporated in the \cref{def:surrogate,thm:opt-round} (i.e. $h_\A(p) \ge \alpha + \beta h_\B(p)$). This result is proved using the MinMax theorem. For the special case when $\ell_\B = \ell_{\text{0-1}}$, we present an alternate proof with an explicit rounding. We will also show that simple projection onto $\Delta_k$ is an optimal rounding rule for $\ell_{\mathrm{CS}}$.

With this characterization, by calculating the entropy functions and coefficients $\beta$ for loss functions, we can compare our results for those losses for the binary and multiclass case, as follows:

\begin{itemize}
    \item $\ell_{\text{weak}}$ is not a surrogate loss for $\ell_{\text{0-1}}$ unless $k=2$, in which case it matches the hinge loss. Thus, for $k >2$, one cannot conclude the hardness of minimizing $\ell_{\text{weak}}$ from hardness for $\ell_{\text{0-1}}$. Indeed, for a deterministic labeling function, the learner which is able to eliminate just $1$ in $k$ labels but then guesses at random will have $0$ loss for $\ell_{\text{weak}}$, but loss $(k-2)/(k-1)$ for $\ell_{\text{0-1}}$. 
    
    \item $\ell_{\mathrm{strong}}$ is a surrogate for $\ell_{\text{0-1}}$ with coefficient $\beta_{\mathrm{strong}} = 4/k$.  In particular, $\delta$-hardness under $\ell_{\text{0-1}}$ implies $4\delta/k$-hardness under $\ell_{\text{strong}}$, and hence a hardcore channel whose noisy labels are at total variation distance at most $1 - 4\delta/k$. When $k=2$, in fact $h_{\mathrm{strong}}(p) \ge 1/2 + h_{\text{0-1}}(p)$, from which $\delta$-hardness under $\ell_{\text{0-1}}$ implies $(1/2 + \delta)$-hardness under $\ell_{\text{strong}}$, and hence a channel whose noisy labels are at distance $1/2 - \delta$, as expected (and shown in Figure \ref{fig:random_outside_S}).

    \item $\ell_{\mathrm{CS}}$ is a surrogate loss with $\beta_{\mathrm{CS}} = k/(k-1)$. Hence $\delta$-hardness under $\ell_{\text{0-1}}$ implies a $\C$-indistinguishable model at TV distance at least  $k\delta/(k-1)$. This is  $2\delta$-distance when $k=2$, where for a Boolean function, such a model can be obtained from a $2\delta$-dense hardcore set $S$ by complementing labels on $S$.
\end{itemize}
Thus we see that the familiar $2\delta$ for the binary setting generalizes differently to larger $k$ for differing loss functions. 

We remark that these are not ``if and only if'' statements, as additional scaling factors are incurred by converting back to $\ell_{\text{0-1}}$ hardness. For example, a $\delta$-distance indistinguishable model only implies $\delta/2$-hardness for $\ell_{\text{0-1}}$, not $(1 - 1/k)\delta$-hardness for $\ell_{\text{0-1}}$, since all the mass at a label $y \in \Y$ may have simply been moved to a different $\tilde{y} \in \Y$, rather than spread out evenly among all labels.

\subsection{Universal Masking Channels From Multicalibration}
\label{sec:intro-universal}

Although \cref{thm:channels} gives an \emph{exact} quantitative relationship between average-case hardness and masking channel fidelity, for each hardness notion $g$ and each associated fidelity notion $h$, the optimal channel $\ch$ may differ depending on the choice of $h$. What if we want a \emph{single} masking channel that witnesses the hardness of $f$ against $\C$ for \emph{all} fidelity functions $h$ simultaneously?

This question of the existence of a \emph{universal} channel brings to mind recent results in learning theory regarding the existence of \emph{omnipredictors}, which are predictive models that outperform a benchmark class $\C$ with respect to a wide range of loss functions $g$ simultaneously, rather than being tailored to one specific such $g$. Perhaps surprisingly, it is known that such omnipredictors indeed exist for arbitrary families of loss functions satisfying mild conditions \cite{omnipredictors,gopalan2023lossoi}. In light of the duality between learning and masking channels, the question posed above seeks the ``dual'' channel  corresponding to an omnipredictor.

We show that even for an entire collection of loss-fidelity pairs $g,h$, there exists a single \emph{universal} model $p : \X \to \Delta_k$ of the conditional label distribution that is not only an omnipredictor for the losses $g$ in the class, but also yields approximately optimal-fidelity masking channels for all $h$ in the family, provided that we slightly enlarge $\C$. Put differently, the action used to minimize $g$ and the channel used to maximize $h$ will only depend on $x$ through the model's prediction $p(x)$. To state this formally, we will express the chosen action as $\hat c_g(p(x))$, where $\hat c_g$ is the \emph{lowest-loss response} to $p(x)$ with respect to $g$, and we will express the chosen channel as $\widehat\ch_h(p(x))$, where $\widehat\ch_h(p(x))$ is the \emph{highest-fidelity response} to $p(x)$ with respect to $h$. We define these best responses as follows:

\begin{definition}[
    name = Lowest-Loss Response,
    label = def:loss-response,
    store = def:loss-response
]
    Given a function $g : \Y \times \R^k \to \eR$, a distribution $z \in \Delta_k$, and a label $\rn{y} \sim z$, a \emph{lowest-loss response} to $z$ with respect to $g$ is any
    \[
        \hat c_g(z) \in \argmin_{c(z) \in \R^k} \E[g(\rn{y}, c(z))] \quad\text{s.t.}\quad \langle c(z), \vec{1} \rangle = 0,
    \]
    assuming any exist. We instead say $\hat c_g$ is \emph{$\gamma$-optimal} if it achieves loss within $\gamma$ of the infimum.
\end{definition}
This lowest-loss response function plays a prominent role in the literature on omniprediction \cite{omnipredictors, gopalan2023lossoi}, where it is used to post-process predictions for a specific loss.

\begin{definition}[
    name = Highest-Fidelity Response,
    label = def:fidelity-response,
    store = def:fidelity-response
]
    Given a fidelity function $h : \Y \times \Delta_k \to \eR$, a distribution $z \in \Delta_k$, and a label $\rn{y} \sim z$, a \emph{best-fidelity response} to $z$ with respect to $h$ is any
    \[
        \widehat\ch_h(z) \in \argmax_{W(z) \in \CSM} \E[h(\rn{y}, W(z)\rn{y})] \quad\text{s.t.}\quad W(z)z=\frac{\vec{1}}{k},
    \]
    assuming any exist. We instead say $\widehat\ch_h$ is \emph{$\gamma$-optimal} if it achieves fidelity within $\gamma$ of the maximum.
\end{definition}
The space of channels we optimize over in this definition maps $z$ to the uniform distribution, as in Example \ref{ex:mask}. Thus high-fidelity responses generalize the re-balancing functions on the level sets of a predictor, which are used in the proof of the hardcore lemma in \cite{TrevisanTV09, casacuberta2024complexity, casacuberta2025global}. 

Better yet, if each fidelity function $h(y, z)$ in the collection is a \emph{nondecreasing} function of $\langle y, z \rangle$, as the ones in \cref{ex:fid-flip,ex:fid-ce}, then there not only exists a universal model $p$, but there also exists a universal highest-fidelity post-processing $\widehat\ch = \widehat\ch_h$ of $p$ that works for \emph{all} fidelity functions $h$ in the collection. We call such fidelity functions \emph{monotone}. Besides monotone fidelity functions, we also consider the natural class of \emph{monotone mixing} fidelity functions, such as the one in \cref{ex:fid-mix}, which constrain the noisy label distribution $z$ to be a mixture of the original label $y$ and the uniform distribution:

\begin{definition}[Monotone Fidelity]
\label{def:mono-fidelity}
    Let $\Y = \{e_1, \ldots, e_k\}$. We say a fidelity function $h : \Y \times \Delta_k \to \eR$ is \emph{monotone} if there is a nondecreasing function $h_0 : \R \to \eR$ such that
    \[
        h(y, z) = \begin{cases}
            h_0(\langle y, z \rangle) &\text{if $z \in \Delta_k$,} \\
            -\infty &\text{otherwise.}
        \end{cases}
    \]
    We instead say that $h$ is \emph{monotone mixing} if for some such $h_0$,
    \[
        h(y, z) = \begin{cases}
            h_0(\mu) &\text{if $z = \mu y + (1-\mu)(\vec{1}/k) \in \Delta_k$ for some $\mu \in [0, 1]$,} \\
            -\infty &\text{otherwise.}
        \end{cases}
    \]
    Although monotone mixing fidelity functions might appear to be a subset of monotone fidelity functions, this is not the case because of their differing effective domains.
\end{definition}

In the following theorem statement, there are three parts. First, part~\ref{item:universal-informal-feasibility} asserts that the channels are indeed ``valid'' or ``feasible'', in the sense that they lower bound the loss of the class $\C$ for $f$. Next, part~\ref{item:universal-informal-optimality} asserts that the bound from part~\ref{item:universal-informal-feasibility} is nearly tight, in the sense that some circuit of ``complexity'' not much higher than that of the complexity class $\C$ indeed achieves similar loss. Finally, part~\ref{item:universal-informal-postproc} asserts that a single masking channel works for all monotone fidelity functions, or for all monotone mixing fidelity functions.

\begin{theorem}[Universal Masking Channels, Informal Version of \Cref{thm:universal-properties,thm:universal-existence,thm:universal-monotone}]
\label{thm:universal-informal}
For all finite $\X$ and $\Y = [k]$, distributions over $(\rn{x}, \rn{y}) \in \X \times \Y$, finite $\C \subseteq \{\X \to \R^k\}$, and $\eps > 0$, and a family $\F$ of tuples $(g, h, \hat c_g, \widehat\ch_h)$, where $g,h$ are convex-concave conjugates and $\hat c_g, \widehat\ch_h$ are best responses, there exists a model $p : \X \to \Delta_k$ of ``complexity'' $O(\log(k)/\eps^2)$ times that of the functions in $\C$ and $\F$,\footnote{In this informal statement, we have omitted an important assumption regarding boundedness of the functions in $\C$ and $\F$. For details, see the formal statement in \cref{thm:universal-existence}.} such that the following properties hold for all $(g,h,\hat c_g, \widehat\ch_h) \in \F$:

\begin{enumerate}[(a)]
    \item \label{item:universal-informal-feasibility} \textbf{(Hardness $\ge$ Average Channel Fidelity)} The channel $\ch_{p} = \widehat\ch_h \circ p$ obtained from post-processing the model $p$ certifies hardness against $\C$, in the sense that for all $c \in \C$,
    \[
        \E[g(\rn{y}, \pi(c(\rn{x})))] \ge \E[h(\rn{y}, W_{p}(\rn{x})\rn{y})] - O(\eps).
    \]
    \item \label{item:universal-informal-optimality} \textbf{(Hardness $\le$ Average Channel Fidelity)} The action $c_p = \hat c_g \circ p$ derived from post-processing the model $p$ is nearly tight for the bound from part~\ref{item:universal-informal-feasibility}:
    \[
        \E[g(\rn{y}, c_p(\rn{x}))] \le \E[h(\rn{y}, W_p(\rn{x})\rn{y})] + O(\eps).
    \]
    Note that even if $c_p \notin \C$, its ``complexity'' is at most $O(\log(k)/\eps^2)$ times that of $\C$.
    \item \label{item:universal-informal-postproc} \textbf{(Universal Masking Channel)} A single $\widehat\ch = \widehat\ch_{h}$ works for all monotone fidelity functions. A (different) single $\widehat\ch = \widehat\ch_{h}$ works for all monotone mixing fidelity functions.
\end{enumerate}
\end{theorem}

The key technical tool underlying the proof of \cref{thm:universal-informal}, which we present in \cref{sec:universal}, is \emph{multicalibration} \cite{hkrr}. Multicalibration is an enhanced notion of computational indistinguishability that has been used both in omniprediction \cite{omnipredictors,gopalan2023lossoi} and in recent new proofs of the traditional hardcore lemma  \cite{casacuberta2024complexity,casacuberta2025global}. A model $p : \X \to \Delta_k$ of $\rn{y}|\rn{x}$ is said to be \emph{$(\C, \eps)$-multicalibrated} if the level sets of $p$ partition $\X$ into regions where the real label $\rn{y}$ is indistinguishable from a simulated label distributed according to $p(\rn{x})$. More formally,
\[
    \E\Bigl[\max_{c \in \C}\bigabs{\E[\langle c(\rn{x}), \, \rn{y} - p(\rn{x})\rangle \mid p(\rn{x})]}\Bigr] \le \eps.
\]
It is known that regardless of the complexity of the conditional distribution $\rn{y}|\rn{x}$, multicalibrated models always exist with complexity comparable to that of $\C$ \cite{hkrr}, and the complexity gap can be shrunk considerably in the cases that we only require indistinguishability on certain weighted combinations of the level sets of $p$, as opposed to all level sets \cite{gopalan2022low}.

\begin{figure}[t]
    \centering
    \begin{subfigure}[t]{0.45\linewidth}
        \centering
        \includegraphics[width=\linewidth]
          {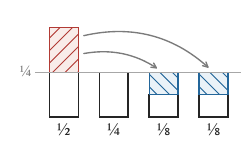}
        \caption{Move mass from above $1/k$ to below $1/k$}
        \label{fig:excess-redistribution}
    \end{subfigure}\hfil
    \begin{subfigure}[t]{0.45\linewidth}
        \centering
        \includegraphics[width=\linewidth]
          {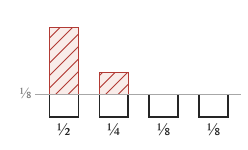}
        \caption{Uniformly redistribute mass above $\min_i z_i$}
        \label{fig:uniform-redistribution}
    \end{subfigure}
    \caption{Multicalibration-based constructions of the universal noisy channel ($k=4$).}
    \label{fig:probability-redistribution}
\end{figure}

Given a multicalibrated model, it is not too hard to see how to construct an optimal universal masking channel for monotone or monotone mixing fidelity functions. Indeed, given a label distribution $z \in \Delta_k$, consider the $z$-level set of the model $p : \X \to \Delta_k$, which we denote by $p^{-1}(z) \subseteq \X$. On this level set, we know that each label $\rn{y}$ is indistinguishable from a fresh sample $z$. Therefore, at least intuitively, any noisy channel that redistributes the mass of $z$ into the uniform distribution should also map the label $\rn{y}$ to a noisy label $\tilde{\rn{y}}$ that is indistinguishable from uniform. Among such channels, it is clear that the highest fidelity (least noisy) choice is the channel that redistributes mass from the label set $S_{z}^{\text{high}} = \{i \in [k] \mid z_i > 1/k\}$ to the label set $S_{z}^{\text{low}} = \{i \in [k] \mid z_i < 1/k\}$. This construction of the universal channel is illustrated in Figure~\ref{fig:excess-redistribution}.

For monotone mixing fidelity functions, which only allow the channel to redistribute mass uniformly, the construction instead uniformly redistributes excess mass from the label set $S^{\neq\text{min}}_z = \{i \in [k] \mid z_i > \min_j z_j\}$. This alternate construction is illustrated in Figure~\ref{fig:uniform-redistribution}. In fact, both constructions can be seen as multiclass generalizations of a previously known multicalibration-based construction of a hardcore measure in the binary 0-1 loss setting \cite{casacuberta2024complexity,casacuberta2025global}. We show in \cref{sec:universal-binary} how to recover this result from our new one.

\begin{remark}
    The formal version of \cref{thm:universal-informal} in \cref{sec:universal} is stronger than the informal version sketched here. While our informal discussion uses multicalibration, the full version will only require \emph{calibration} \cite{Dawid1982TheWB} and weighted \emph{multiaccuracy} \cite{TrevisanTV09,hkrr}. These conditions are more efficiently achieved than multicalibration, improving the complexity of $p$ relative to $\C$.
\end{remark}

\subsection{Characterizing Pseudoentropy via MaxEnt}
\label{sec:intro-maxent}

In this section, we  discuss the relationship between our results on masking channels and related work on \emph{pseudoentropy} \cite{vadhan2012pseudoentropy,zheng2014thesis,hu2025generalized}. We show that a tight learning-theoretic characterization of pseudoentropy can be derived using the \emph{MaxEnt} algorithm \cite{berger1996maxent,lafferty2000bregman,dudik2007maxent,altun2006maxent}.

\begin{figure}
    \centering
    \includegraphics[width=0.7\linewidth]{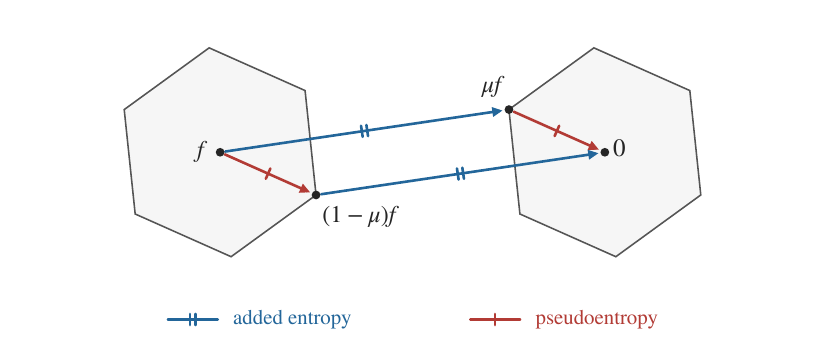}
    \caption{Symmetry between \cref{thm:channels} (Noisy Channels) and \cref{thm:maxent} (MaxEnt) for binary 0-1 loss. The polytopes are $\C$-indistinguishability regions, one centered at the original $f$ and another at pure noise $0$. If $f$ ($\{\pm 1\}$-valued) is $\delta$-hard for $\C$ and  if $\mu$ ($[0, 1]$-valued) is a $2\delta$-dense hardcore measure, the high-entropy model is $(1 - \mu)f \approx_\C f$ and the noisy channel output is $\mu f \approx_\C 0$.}
    \label{fig:maxent-vs-hardcore}
\end{figure}

Much like our high fidelity masking channels, the property of having high \emph{pseudoentropy} \cite{hastad1999hill}, generalized in \cite{vadhan2012pseudoentropy,zheng2014thesis,hu2025generalized}, is a hardness-certifying relaxation of the hardcore property. This property states that $f : \X \to \{\pm 1\}$ itself is $\C$-indistinguishable from a model $p : \X \to [-1, +1]$ that is highly random. In this most basic setting, ``highly random'' means being $(1/2-\delta)$-close in statistical distance to pure random noise. Equivalently, the model $p$ has high \emph{$\ell_{\text{0-1}}$ entropy}: $
    \E[h_{\text{0-1}}(p(\rn{x}))] \ge \delta$,
where \(
    h_{\text{0-1}}(v) = (1 - |v|)/2
\)
(see \cref{def:entropy}). As with the masking channels property, it is easy to check that high pseudoentropy certifies hardness. Also, randomizing labels on the hardcore measure yields the high-entropy model $p(x) = (1 - \mu(x))f(x)$. There is a symmetry between these two approaches to characterizing average-case hardness, which Figure~\ref{fig:maxent-vs-hardcore} depicts in the most basic setting.

As shown in \cite{vadhan2012pseudoentropy,zheng2014thesis}, having high pseudoentropy, much like our novel noisy channel property, extends naturally to the multiclass setting and to essentially arbitrary concave ``entropy'' functions $h : \Delta_k \to \R$, such as the 0-1 entropy function $1 - \max_i p_i$, the Shannon entropy $\sum_i p_i\log(1/p_i)$, and many more. They show that for any reasonably well-behaved entropy function $h$ and associated \emph{GLM loss} function $\ell$, the best loss attainable by circuits in a class $\C'$ related to $\C$ is approximately equal to the maximum entropy of a $\C$-indistinguishable model.\footnote{Technically, the equivalence stated in \cite{zheng2014thesis} is not between GLM loss minimization and entropy maximization, but rather between minimization of a \emph{Bregman divergence} and maximization of a certain entropy \emph{gap}. These two formulations turn out to be exactly equivalent.} 

The elegant recent work of Hu and Vadhan \cite{hu2025generalized} shows that in fact, even for an entire collection of entropy functions $h$ and associated GLM loss functions, there exists a \emph{single} universal model $p$ with complexity comparable to that of $\C$, whose $h$-entropy certifies the hardness of $f$ for the class $\C$, and whose own loss on $f$ is close to the certified lower bound. 

Our new \cref{thm:universal-informal} on universal masking channels is inspired by their result and this symmetry between masking channels and pseudoentropy. But this raises the question of whether there is a parallel to \cref{thm:channels}. In other words,
is there a learning problem that tightly characterizes pseudoentropy?

We show that the answer comes from a line of work in the natural language processing on the \emph{MaxEnt} algorithm \cite{berger1996maxent,lafferty2000bregman,dudik2007maxent,altun2006maxent}. The MaxEnt algorithm attempts to learn a probability distribution of the highest possible entropy, subject to a collection of linear constraints imposed by observed data points. In our setting, we wish to learn $p(x) =\E[\y|\x = x]$ under the linear constraints of $\C$-indistinguishability. The objective is $\E[h(p(\x))]$, the model's average $h$-entropy. The convex dual is known to be the problem of  minimizing a  \emph{generalized linear model (GLM)} loss over $\spn(\C)$, with $\ell_1$ regularization; yielding a learning problem that exactly characterizes pseudoentropy.

\begin{theorem}[
    name = {Characterizing pseudoentropy using MaxEnt \cite{berger1996maxent,lafferty2000bregman,dudik2007maxent,altun2006maxent}},
    label = thm:maxent,
    store = thm:maxent
]
    For all finite $\X$ and $\Y = [k]$, distributions $\D$ over pairs $(\rn{x}, \rn{y}) \in \X \times \Y$, finite $\C \subseteq \{\X \to \R^k\}$ and convex-concave conjugate pairs $g : \R^k \to \R$ and $h : \Delta_k \to \eR$, let $\ell(y, \eta) = g(-\eta) - \langle y, \eta \rangle$ (GLM loss). Then, for all $\eps \ge 0$,\footnote{
    When $\eps \neq 0$ and $\ell$ is bounded below, the infimum on the left-hand side can be replaced with a minimum.}
    \[
        \inf_{\lambda \in \R^\C}\,\E[\ell(\rn{y}, c_\lambda(\rn{x}))] + \eps\|\lambda\|_1
        = \max_{p : \X \to \Delta_k}\, \E[h(p(\rn{x}))] 
            \quad\text{s.t.}\quad\text{$p(\rn{x})$ is $(\C, \eps)$-indistinguishable from $\rn{y}$.}
    \]
\end{theorem}
In \cref{sec:maxent}, we will show that \cref{thm:maxent}, just like our new \cref{thm:channels,thm:universal-informal}, can be derived from the same underlying general framework based on convex duality.

\subsection{Related Work}

Impagliazzo's Hardcore Lemma \cite{impagliazzo1995hard, holenstein2005key} is a cornerstone result in our understanding of the average-case complexity of Boolean functions. It has applications to hardness amplification \cite{impagliazzo1995hard, TrevisanXOR, ODonnell2004Hardness},  the derandomization of {\rm BPP} \cite{ImpagliazzoWigderson}, cryptography \cite{holenstein2005key} and the design of better boosting algorithms \cite{KlivansServedio}. There has been an extensive body of work devoted to optimizing, generalizing and extending this result. While a comprehensive survey of this work is beyond the scope of this section, we identify three broad approaches to proving the hardcore lemma that are relevant to our work.

\paragraph{Boosting/Learning.} As observed by Klivans and Servedio \cite{KlivansServedio}, Impagliazzo's original constructive proof is naturally viewed as a boosting procedure. We assume that the function $f$ is not too hard for dense subsets; that there is a hypothesis $c_S$ that does well for every dense subset $S$. We use boosting to combine these hypotheses into one that does well on average. This approach is further developed in the works of \cite{holenstein2005key, BarakHardtKale, feldman2010boosting}. Better boosting algorithms yield improvements in various parameters of the hardcore lemma, and more recently in works such as \cite{brukhim2020online,brukhim2023simple,bressan2025dice}.

\paragraph{MinMax.} Impagliazzo  presents an alternate proof of the hardcore lemma by Nisan using the MinMax theorem \cite{impagliazzo1995hard}.  We consider a two-player zero-sum game between a learner who makes predictions using hypotheses $c \in \C$ and an adversary who uses dense subsets $S$, the payoff is how well $c$ can predict $f$ on $S$. Depending on the value of the game, either the function is easy on average, or there is a dense hardcore measure. This approach is used by \cite{holenstein2005key, TrevisanTV09, vadhan2012pseudoentropy, vadhan2013uniform}. 

\paragraph{Regularity.} This approach was initiated in the seminal work of Trevisan, Tulsiani and Vadhan \cite{TrevisanTV09}. We use a randomized function  $p:\X \to [-1,1]$ which is  a {\em regular} model of the Boolean function $f:\X \to \{\pm 1\}$, meaning it is indistinguishable from $f$ to $\C$.\footnote{The term ``regularity'' has its roots in \emph{graph regularity lemmas} in combinatorics \cite{szemeredi1975regular,frieze1996regularity}.} We then explicitly reweight the input $x$ using the pair $(p(x), f(x))$ to get a hardcore measure. The notion of regularity in \cite{TrevisanTV09} corresponds to multiaccuracy \cite{hkrr}. More recent works \cite{casacuberta2024complexity, casacuberta2025global, hu2025generalized}  derive stronger results using stronger regularity notions, such as multicalibration or calibrated multiaccuracy which originated in the multigroup fairness literature \cite{hkrr, omnipredictors, gopalan2023lossoi}. We discuss these in more detail shortly.  

\medskip

\noindent Our work builds on key ideas from each of these lines of work, and generalizes them substantially. The use of Fenchel duality considerably generalizes the use of the MinMax Theorem for 2 player games, which is implied by it \cite{Bertsekas2015}. While learning and hardness have long been understood to be two sides of the same coin, our formulation reveals them to be convex duals. Strong notions of  regularity are crucial in our results on universal masking channels, which simultaneously certify hardness for a number of learning tasks, which we view  as {\em dual objects to omnipredictors}. 

Having discussed the techniques used to prove the hardcore lemma, we now discuss extensions of the original hardcore lemma which are particularly relevant to our work.

\paragraph{Beyond 0-1 Loss.} Much of the initial work on the hardcore lemma focuses on 0-1 loss and min-entropy. The work of Vadhan and Zheng considered cross-entropy loss and Shannon entropy, motivated by simpler constructions of pseudorandom generators from one-way functions \cite{vadhan2012pseudoentropy,vadhan2013uniform}.  Zheng's thesis considers more general entropy functions and Bregman divergences \cite{zheng2014thesis}. These works view the hardcore lemma as explaining hardness through pseudoentropy; given a distribution $(\rn{x}, f(\rn{x}))$, the aim is to construct a $\C$-indistinguishable distribution $(\rn{x}, \rn{y'})$ that has high conditional entropy $\E[H(\rn{y'}|\rn{x})]$. They characterize pseudoentropy by the optimal Bregman divergence achievable when sampling $\rn{y}|\rn{x}$ using an efficient sampler. By standard characterizations of proper losses \cite{savage1971elicitation}, this is equivalent to minimizing a proper loss on the distribution $(\rn{x}, \rn{y})$.

\paragraph{Multiple Losses at Once.} The work of \cite{omnipredictors} introduced the notion of an $(\mathcal{L}, \C)$-omnipredictor, which is a single universal predictor that can perform competitively with the best hypothesis from a class $\C$, for any loss in a family $\mathcal{L}$ of loss functions. The best hypothesis in $\C$ can be specific to the loss function, but the omnipredictor is loss-oblivious. The elegant recent work of Hu and Vadhan \cite{hu2025generalized} gives an analogous statement from the hardness perspective: they show that a suitable predictor $p$ yields a joint distribution $(\rn{x}, \rn{y} \sim p(\rn{x}))$ that has optimal pseudoentropy for a large family of entropy functions. The conditions on $p$ (calibrated multiaccuracy) and on the  entropy/loss functions (proper loss) exactly match those needed for loss outcome indistinguishability (loss OI) for proper loss functions \cite{gopalan2023lossoi}, which is a strong form of omniprediction. Our construction of universal masking channels is inspired by their work. 

\paragraph{Surrogate Losses.}
Surrogate losses are used in machine learning as a way to come up with tractable convex substitutes for non-convex losses like 0-1 loss \cite{BartlettJordanMcAuliffe2006, TewariBartlett2007}. Indeed, many commonly used convex losses like hinge loss, $\ell_1$ loss and logistic loss are known to give upper bounds on the 0-1 loss \cite{BartlettJordanMcAuliffe2006}. The tight factor of $2$ for $\ell_1$ and hinge loss is crucial to optimality in the learning algorithm of \cite{KKMS08}. Other works that give bounds for surrogate losses for 0-1 loss in the multiclass setting include  \cite{VanDerHoeven2020, VanDerHoevenFuscoCesaBianchi2021}. Theorem \ref{thm:opt-round} which gives a tight characterization of the optimal rounding factor for any pair of losses via entropy appears to be new.

\subsection{AI Methodology}

ChatGPT 5.6 and 6 were used for several brainstorming discussions, to speed up the calculation of numerous convex conjugates, dual programs, and entropy functions, to determine the rounding schemes presented in \cref{thm:opt-round,thm:cs-round} and the associated entropy functions and surrogate constants, to determine the precise technical conditions needed for strong duality and the attainment of optimizers in our application of convex duality, to search for and understand relevant related work, and to generate figures from screenshots of hand-drawn sketches. All other theorems were formulated and proved without substantial AI assistance. None of the writing in this paper (exposition, definitions, theorems, proofs, etc.) was done by an AI model.

\section{Preliminaries}

In this section, we review the relevant background material on convex analysis (\cref{sec:prelim-convex}) and multicalibration (\cref{sec:prelim-multical}) needed to prove our main results. We especially encourage readers who have not previously seen Fenchel duality to read \cref{thm:fenchel} and the subsequent remarks, as there are some differences from the more familiar notion of linear programming duality.

\subsection{Convex Analysis}
\label{sec:prelim-convex}

In this section, we briefly describe the elements of convex analysis that we require to prove our general theorem on average-case hardness. For simplicity, we restrict attention to the finite-dimensional case, which is the only case that we will require in this paper. We assume the reader is already familiar with the definitions of convex sets and convex/concave functions. The following definitions are all more or less standard (see, for example, \cite{rockafellar1970convex}).

\begin{definition}[Extended Real Numbers]
    Let $\eR = \R \cup \{\pm \infty\}$ denote the extended real numbers. The \emph{effective domain} of an extended-real-valued function $f : \R^n \to \eR$ is the set
    \[
        \dom f = \{x \in \R^n \mid f(x) \in \R\}.
    \]
\end{definition}

\begin{definition}[Closed Proper Convex/Concave Functions]
\label{def:convex}
    We say that $g : \R^n \to \eR$ is a \emph{closed proper convex function} if it is lower semicontinuous (i.e. has closed sublevel sets $g^{-1}((-\infty, v])$ for each $v \in \R$), if it is proper (i.e. never outputs $-\infty$, does not always output $+\infty$), and if it is convex over its effective domain, which must be a convex set. If $g : S \to \eR$ is defined only on a subset $S \subseteq \R^n$, then whether or not $g$ is a closed proper convex function is judged from its extension
    \[
        \bar{g}(x) = \begin{cases}
            g(x) &\text{if $x \in S$,}\\
            +\infty &\text{if $x \notin S$.}
        \end{cases}
    \]
    We say that $h : S \to \eR$ is a \emph{closed proper concave function} if $-h$ is a closed proper convex function.
\end{definition}

\begin{example}[Convex Indicator]
\label{def:convex-indicator}
    If $S \subseteq \R^k$ is a closed nonempty convex set, then its \emph{convex indicator} $I_S : \R^k \to \eR$, defined as follows, is a closed proper convex function:
    \[
        I_S(x) = \begin{cases}
            0 &\text{if $x \in S$,}\\
            +\infty &\text{if $x \notin S$.}
        \end{cases}
    \]
\end{example}

\begin{definition}[Subgradients]
    We say that a vector $v \in \R^n$ is a \emph{subgradient} of a function $f : \R^n \to \eR$ at $a \in \R^n$ if for all $b \in \R^n$,
    \[
        f(b) \ge f(a) + \langle v, b - a \rangle.
    \]
    In general, subgradients need not exist or be unique. The set of all subgradients of $f$ at $a$ is called the \emph{subdifferential} at $a$, and it is denoted by $\partial f(a)$. (It is known that if $f$ is convex, then $\partial f(a) = \{\nabla f(a)\}$ at each point $a$ where $f$ is differentiable.)
\end{definition}
\begin{definition}[Convex Conjugates]
\label{def:conj-0}
    We say that two functions $f,g : \R^n \to \eR$ are \emph{convex conjugates} if they are both closed proper convex functions and
    \begin{align*}
        f(a) = \sup_b\, \langle a, b \rangle - g(b),\\
        g(b) = \sup_a\, \langle a, b \rangle - f(a).
    \end{align*}
    (It is known that every closed proper convex function $f$ has a unique convex conjugate $g$, and that applying this transformation twice recovers the original $f$.)
\end{definition}
\begin{definition}[Convex-Concave Conjugates]
\label{def:conj}
    We say that $g,h : \R^n \to \eR$ are \emph{convex-concave conjugates} if $g$ is a closed proper convex function, $h$ is a closed proper concave function, and the functions $a \mapsto g(-a)$ and $b \mapsto -h(b)$ are convex conjugates as per \cref{def:conj-0}. Equivalently,
    \begin{align*}
        g(a) &= \sup_b\, h(b) - \langle a, b \rangle,\\
        h(b) &= \inf_a\, g(a) + \langle a, b \rangle.
    \end{align*}
\end{definition}
\begin{definition}[Convex-Concave Conjugates With Two Inputs]
\label{def:conj-2}
    Given finite $\Y = [k]$, we say that the two-input functions $g, h : \Y \times \R^n \to \eR$ are convex-concave conjugates if the single-input functions $g(y, \cdot)$ and $h(y, \cdot)$ are convex-concave conjugates as per \cref{def:conj} for all $y \in \Y$.
\end{definition}

\begin{definition}[Duality Map]
\label{def:dual-map}
    Given convex conjugates $f,g : \R^n \to \eR$, we say that points $a,b \in \R^n$ are \emph{dual} to each other if they make equality hold in the following inequality, called the \emph{Fenchel-Young inequality}, which holds for all $a,b \in \R^n$:
    \[
        f(a) + g(b) \ge \langle a,b \rangle.
    \]
    (When $f$ and $g$ are differentiable, it is known that $a,b$ are dual to each other iff $a = \nabla g(b)$ iff $b = \nabla f(a)$. In general, $a,b$ are dual to each other iff $a \in \partial g(b)$ iff $b \in \partial f(a)$.) Similarly, given a convex-concave conjugate pair $g,h$, we say that $a,b$ are dual to each other if the following inequality holds with equality:
    \[
        g(a) \ge h(b) - \langle a, b \rangle.
    \]
\end{definition}

\begin{theorem}[Fenchel's Duality Theorem, Special Case \cite{rockafellar1970convex}]
\label{thm:fenchel}
    Let $g_1, h_1 : \R^n \to \eR$ and $g_2,h_2 : \R^n \to \eR$ be two convex-concave conjugate pairs. If at least one of these four functions is finite everywhere (i.e. never outputs $\pm \infty$), then
    \begin{equation}
    \label{eq:fenchel}
        \inf_a\, g_1(a) + g_2(a) = \sup_b\, h_1(b) + h_2(-b).
    \end{equation}
    If the finite function among the four was one of $g_1,g_2$, then the supremum on the right-hand side is achieved by some particular $b^* \in \R^n$. If the finite function among the four was one of $h_1, h_2$, then the infimum on the left-hand side is achieved by some particular $a^* \in \R^n$.
\end{theorem}

\begin{remark}
    To build intuition for \cref{thm:fenchel}, here is a neat ``proof'':
    \begin{align*}
        \inf_a\, \bigl(g_1(a) + g_2(a)\bigr) &= \inf_{a_1,a_2}\, \sup_b\, \bigl(g_1(a_1) + g_2(a_2) + \langle a_1 - a_2, b \rangle \bigr)\\
        &\overset{?}{=} \sup_b \inf_{a_1,a_2}\, \bigl(g_1(a_1) + g_2(a_2) + \langle a_1 - a_2, b \rangle \bigr) \\
        &= \sup_b\, \bigl(\inf_{a_1}\, g_1(a_1) + \langle a_1, b \rangle\bigr) + \bigl(\inf_{a_2}\, g_2(a_2) - \langle a_2, b \rangle\bigr) \\
        &= \sup_b\, h_1(b) + h_2(-b).
    \end{align*}
    The preceding calculation is not entirely rigorous, since it swaps the inf and sup without proper justification. Nevertheless, it conveys the main idea. There also exist simple and fully rigorous proofs of \cref{thm:fenchel} using the separating hyperplane theorem.
\end{remark}

\begin{remark}
    It is clear that $a,b$ achieve equality in \eqref{eq:fenchel} (i.e. are a ``saddle point'') iff they are dual for $g_1, h_1$ and $a,-b$ are dual for $g_2, h_2$. Indeed, by the Fenchel-Young inequality, we have
    \begin{align*}
        g_1(a) + g_2(a) \ge \bigl(h_1(b) - \langle a, b \rangle\bigr) + \bigl(h_2(-b) + \langle a, b \rangle\bigr) = h_1(b) + h_2(-b),
    \end{align*}
    with equality iff the two duality conditions hold (see \cref{def:dual-map}).
\end{remark}

\begin{remark}
\label{rmk:fenchel-weakening}
    The finiteness assumption in our statement of \cref{thm:fenchel} is actually far stronger than necessary, not only for equating the primal value $p = \inf_a g_1(a) + g_2(a)$ with dual value $d = \sup_b h_1(b) + h_2(-b)$, but also for proving that these values are attained when finite. In fact, to prove that there exists $b^* \in \R^n$ with $h_1(b^*) + h_2(-b^*) = p = d$, we only require there to be a non-vertical hyperplane that weakly separates the disjoint convex sets $\{(a, t) \mid t > g_1(a)\}$ and $\{(a, t) \mid t < p - g_2(a)\}$. A weakly separating hyperplane between two disjoint convex sets always exists by the separating hyperplane theorem, and assuming that $g_1$ or $g_2$ is finite everywhere is just one of many convenient ways to force this hyperplane to be non-vertical.
\end{remark}

\subsection{Hinge Loss Rounding}
\label{sec:surrogate}

We now discuss the known relationship between 0-1 loss, the binary hinge loss $\ell(y, a) = (1 - ay)_+$, and the Crammer-Singer multiclass hinge loss \cite{crammer2001multiclass}
\[
    \ell_{\mathrm{CS}}(y, a) = (1 - a_y + \max_{j \neq y} a_j)_+.
\]

\begin{remark}[Hinge Loss Rounding {\cite[Section 3.2]{KKMS08}}] 
\label{rmk:01-vs-hinge}
Given $a \in \R$, let $\tilde{a} = \mathrm{proj}_{[-1, +1]}(a)$ be its projection onto the interval $[-1, +1]$. If we sample a random threshold $\rn{t} \sim [-1, +1]$, then
\[
    \ell_{\mathrm{hinge}}(y, a) \ge \ell_{\mathrm{hinge}}(y, \tilde{a}) = \abs{y - \tilde{a}} = 2\cdot\E_{\rn{t}}[\ell_{\text{0-1}}(y, \sign(a - \rn{t}))].
\]
Thus, a $2\delta$-upper bound on hinge loss implies a $\delta$-upper bound on 0-1 loss.
\end{remark}

\begin{remark}[Crammer-Singer Rounding \cite{crammer2001multiclass}]
\label{rmk:cs-hinge-01}
    Given $a \in \R^k$, let $j^\star \in \argmax_j a_j$. Then,
    \[
        \bm{1}[y \neq j^\star] \le \ell_{\mathrm{CS}}(y, a).
    \]Thus, a $\delta$-upper bound on Crammer-Singer hinge loss implies a $\delta$-upper bound on 0-1 loss.
\end{remark}

\subsection{Multicalibration}
\label{sec:prelim-multical}

Here, we state the precise forms of computational indistinguishability that we require for the proof of our universal multiclass hardcore lemma. There are two definitions that we need: calibration \cite{Dawid1982TheWB} and multiaccuracy \cite{hkrr,kearns2018gerrymandering}.

We begin with the definition of weight-restricted calibration. This condition requires the error of the model $p : \X \to \Delta_k$ to be approximately uncorrelated with the model's predictions themselves, even after post-processing by some appropriately bounded weight functions in a prespecified collection $\W$. We remark that the notational similarity of $\W$ (the weight function class) and $\ch$ (our noisy channel) is not intentional but nevertheless convenient since the former will be largely determined by the latter.

\begin{definition}[Weighted Calibration \cite{Dawid1982TheWB,gopalan2022low}]
\label{def:calibration}
    Given finite $\X$ and $\Y = [k]$, a distribution over $(\rn{x}, \rn{y}) \in \X \times \Y$, and $\W \subseteq \{\Delta_k \to \R^k\}$, we say that $p : \X \to \Delta_k$ is \emph{$(\W, \eps)$-calibrated} if for all $w \in \W$,
    \[
        \bigabs{\E[\langle w(p(\rn{x})), \, \rn{y} - p(\rn{x})\rangle]} \le \eps.
    \]
\end{definition}

Next, we define multiaccuracy and its weighted variant. Intuitively, multiaccuracy requires the error of $p$ to be approximately uncorrelated with each function in a prespecified collection $\C$. The weighted version also includes a factor depending on the model's predictions themselves, as in the definition of a calibration weight function.

\begin{definition}[Multiaccuracy \cite{hkrr,kearns2018gerrymandering}]
\label{def:multiaccuracy}
    Given finite $\X$ and $\Y = [k]$, a distribution over $(\rn{x}, \rn{y}) \in \X \times \Y$, and a collection $\C \subseteq \{\X \to \R^k\}$, we say that $p : \X \to \Delta_k$ is \emph{$(\C, \eps)$-multiaccurate } if for all $c \in \C$,
    \[
        \bigabs{\E[\langle c(\rn{x}), \, \rn{y} - p(\rn{x})\rangle]} \le \eps.
    \]
\end{definition}
\begin{definition}[Weighted Multiaccuracy \cite{gopalan2022low,casacuberta2025global}]
\label{def:weighted-multiaccuracy}
    In the setting of \cref{def:multiaccuracy}, given a weight function $w : \Delta_k \to \R^{k \times k}$, we say that $p$ is \emph{$(\C, \eps)$-multiaccurate with respect to $w$} if for all $c \in \C$,
    \[
        \bigabs{\E[\langle w(p(\rn{x})) c(\rn{x}), \, \rn{y} - p(\rn{x})\rangle]} \le \eps.
    \]
    Note that standard $(\C, \eps)$-multiaccuracy corresponds to the case that $w(\cdot) = I_{k \times k}$.
\end{definition}

It is by now well known that for any class $\C$ (e.g. a Boolean circuit complexity class), one can construct a multiaccurate predictor $p$ of ``complexity'' not much greater than that of $\C$ itself. By this, we mean that $p(x)$ can be evaluated by performing simple arithmetic operations on the values $c_1(x), \ldots, c_m(x)$, where $m = O(\log(k)/\eps^2)$ and the functions $c_1, \ldots, c_m$ all belong to the class $\C$. With $k = 2$, this has been known since the work of \cite{TrevisanTV09}, and the fact that the same bound holds even if one incorporates weight functions depending on the model itself in complex ways has been known since \cite{hkrr,dwork2021outcome}.

\begin{theorem}[Outcome Indistinguishability via Multiplicative Updates \cite[Example 5.6]{dwork2023pseudorandomness}]
\label{thm:multicalibration}
For all finite $\X$ and $\Y = [k]$, distributions over $(\rn{x}, \rn{y}) \in \X \times \Y$, collections of functions $\F \subseteq \{\Delta_k \times \X \to [-1, +1]^k\}$, and $\eps > 0$, there exists $p : \X \to \Delta_k$ with the following properties:
\begin{itemize}
    \item \textbf{(Indistinguishability)} For all functions $f \in \F$,
    \[
        \bigabs{\E[\langle f(p(\rn{x}), \rn{x}), \, \rn{y} - p(\rn{x})\rangle]} \le O(\eps).
    \]
    \item \textbf{(Complexity Bound)} The model $p$ has the form $p_m$, where we define $p_0(x) = \vec{1}/k$
    and
    \[
        p_j(x) \propto \exp\Bigl(\eps \sum_{i=1}^{j} f_i(p_{i-1}(x), x)\Bigr),
    \]
    normalized so that $\sum_y p_j(x)_y = 1$, for some $m = O(\log(k)/\eps^2)$ functions $f_1, \ldots, f_m \in \pm \F$.
\end{itemize}
\end{theorem}

\section{A General Framework}
\label{sec:general}

In this section, we present a very general framework, captured by \cref{thm:six}, which we will instantiate in later sections to prove our main results. We will also see in later sections that this framework can be used to derive several of the prior results on pseudoentropy and the MaxEnt algorithm that we discussed in the introduction.

Before stating the result, let us recall from \cref{sec:prelim-convex} the precise statement of our convex-concave conjugacy requirement on the loss function $g$ and fidelity function $h$:
\begin{equation}
\label{eq:general-loss-fidelity-conj}
\begin{split}
    g(y, a) &= \sup_{b \in \R^k}\, h(y, b) - \langle a, b \rangle, \\
    h(y, b) &= \inf_{a \in \R^k}\, g(y, a) + \langle a, b \rangle.
\end{split}
\end{equation}
Since a supremum of linear functions is convex, and since an infimum of linear functions is concave, these conditions imply that $g(y, \cdot)$ and $h(y, \cdot)$ are convex and concave, respectively.  Later, we will also require the effective domain of $h$ in its second input to be contained in $\Delta_k$, but we do not yet need to impose that requirement.

Our general result will be stated not only in terms of $g,h$, but also in terms of a pair of \emph{regularizer} or \emph{constraint} functionals $R,S$. These take as input a function from $\X \times \Y$ to $\R^k$, and they output a value in $\eR$. Intuitively, we will use these functions to impose penalties like the $\eps \|\lambda\|_1$ penalty appearing in \eqref{eq:intro-hardcore} and \cref{thm:channels}, as well as the indistinguishability constraints appearing in \eqref{eq:intro-hinge} and \cref{thm:channels}. We require the functionals $R$ and $S$ to satisfy a conjugacy relationship similar to the one satisfied by $g$ and $h$:
\begin{equation}
\label{eq:general-RS-conj}
\begin{split}
    R(a) &= \sup_{b : \X \times \Y \to \R^k}\, S(b) - \E\bigl[\langle a(\rn{x}, \rn{y}), b(\rn{x}, \rn{y}) \rangle\bigr], \\
    S(b) &= \inf_{a : \X \times \Y \to \R^k}\, R(a) + \E\bigl[\langle a(\rn{x}, \rn{y}), b(\rn{x, \rn{y}}) \rangle\bigr].
\end{split}
\end{equation}
There are a couple key differences between the conjugacy relationships in \eqref{eq:general-loss-fidelity-conj} and \eqref{eq:general-RS-conj}. The first difference is that \eqref{eq:general-loss-fidelity-conj} optimizes over $k$-dimensional vectors $a, b \in \R^k$, but \eqref{eq:general-RS-conj} optimizes over entire \emph{functions} $a, b : \X \times \Y \to \R^k$, which can be viewed as $\abs{\X}\abs{\Y}k$-dimensional vectors. The second difference is that \eqref{eq:general-loss-fidelity-conj} uses the standard inner product in $\R^k$, but \eqref{eq:general-RS-conj} uses an inner product that crucially depends on the given distribution over $(\rn{x}, \rn{y})$ pairs. We now give two examples of $R,S$ pairs of interest. The first is the one that we saw implicitly in the introduction.

\begin{definition}[Indistinguishability Constraint Functionals]
\label{def:hardcore-constraint}
    Given finite $\X$ and $\Y$, a distribution over $(\rn{x}, \rn{y}) \in \X \times \Y$ a finite collection $\C \subseteq \{\X \to \R^k\}$, and $\eps \ge 0$, the \emph{indistinguishability constraint functionals} $R,S$ are defined as follows. For a function $b : \X \times \Y \to \R^k$, the concave functional $S(b)$ enforces the constraint that $b$ is approximately uncorrelated with each $c \in \C$:
    \[
        S(b) = \begin{cases}
            0 &\text{if $\abs{\E[\langle c(\rn{x}), b(\rn{x},\rn{y})\rangle]} \le \eps$ for all $c \in \C$,} \\
            -\infty &\text{otherwise.}
        \end{cases}
    \]
    Next, for $a : \X \times \Y \to \R^k$, the convex functional $R(a)$ checks that $a \in \spn(\C)$ (in particular, $a$ must ignore its $y$-input), and then applies an $\ell_1$ penalty to the coefficients $\lambda$ used to form $a$:
    \[
        R(a) = \inf_{\lambda \in \R^\C} \, \eps \|\lambda\|_1  \text{ such that $\Pr\Bigl[a(\rn{x}, \rn{y}) = \sum_i \lambda_i c_i(\rn{x})\Bigr] = 1$.}
    \]
    Note that $R(a) = +\infty$ if $a \notin \spn(\C)$. The types of these functionals are $R,S : \{\X \times \Y \to \R^k\} \to \eR$.
\end{definition}

\begin{definition}[Indistinguishability Penalty Functionals]
\label{def:hardcore-penalty}
    Given finite $\X$ and $\Y$, a distribution over $(\rn{x}, \rn{y}) \in \X \times \Y$ and a \emph{convex compact} collection $\overline{\C} \subseteq \{\X \to \R^k\}$,\footnote{In particular, $\C$ is not finite whenever it contains two or more elements!} the \emph{indistinguishability penalty functionals} $R,S$ are defined as follows. For a function $b : \X \times \Y \to \R^k$, the concave functional $S(-b)$ penalizes $b$ for being too positively correlated with any $c \in \overline{\C}$:
    \[
        S(-b) = -\max_{c \in \overline{\C}}\, \E[\langle c(\rn{x}), b(\rn{x},\rn{y})\rangle].
    \]
    Note that $S(-b)$ is roughly $0$ when $b$ is uncorrelated with $\C$ but takes on a large negative value when some $b$ is positively correlated with some $c \in \overline{\C}$.\footnote{The reason for focusing on $S(-b)$, as opposed to $S(b)$, will become apparent later when we state \cref{thm:six}.} We also write ``max'' instead of ``sup'' since $\overline{\C}$ is assumed to be compact. Next, for $a : \X \times \Y \to \R^k$, the convex functional $R$ simply checks that $a \in \spn(\C)$ (in particular, $a$ must ignore its $y$-input):
    \[
        R(a) = \begin{cases}
            0 &\text{if $\Pr[a(\rn{x}, \rn{y}) = c(\rn{x})] = 1$ for some $c \in \overline{\C}$,}\\
            +\infty &\text{otherwise.}
        \end{cases}
    \]
    This is closely related to the \emph{convex indicator} $I(\cdot)$ function defined in \cref{def:convex-indicator}. Finally, as in \cref{def:hardcore-constraint}, the types of these functionals are $R,S : \{\X \times \Y \to \R^k\} \to \eR$.
\end{definition}

Our general theorem, which we state and prove below, asserts that for any suitable $g,h,R,S$, minimizing $g$ in expectation over functions $a$ that have been penalized (or constrained) by $R(a)$ is equivalent to maximizing $h$ in expectation over functions $b$ that have been penalized (or constrained) by $S(-b)$. Despite the seemingly abstract form of the theorem, involving six functions $a,b,g,h,R,S$, the proof is actually a surprisingly straightforward application of Fenchel duality (\cref{thm:fenchel}). The theorem statement will inherit some mild technical assumptions from our stated version of Fenchel duality, which, as we explain in \cref{rmk:fenchel-weakening}, can easily be relaxed.

In subsequent sections, we will see even more examples of pairs $g,h$ and pairs $R,S$ that convey the significance of \cref{thm:six}. For example, in \cref{sec:channels}, we apply the general result to prove our main result on channel fidelity, \cref{thm:channels}, and its soft penalty variant. In \cref{sec:universal}, we use it to prove our universal hardcore lemma, \cref{thm:universal-informal}. In \cref{sec:maxent}, we use it to recover MaxEnt duality, \cref{thm:maxent}, as a special case. Finally, in \cref{sec:applications}, we use it to prove several additional interesting and concrete multiclass generalizations of the hardcore lemma.

\begin{theorem}[Six Functions]
\label{thm:six}
    Fix finite sets $\X, \Y$, a distribution $\D$ over $(\rn{x}, \rn{y}) \in \X \times \Y$, functions $g,h : \Y \times \R^k \to \eR$, and functionals $R,S : (\X \times \Y \to \R^k) \to \eR$. Then, under the three technical conditions stated below, strong duality holds:
    \begin{equation}
    \label{thm:six-strong-duality}
        \inf_{a : \X \times \Y \to \R^k}\, \E[g(\rn{y}, a(\rn{x}, \rn{y}))] + R(a) = \sup_{b : \X \times \Y \to \R^k}\, \E[h(\rn{y}, b(\rn{x}, \rn{y}))] + S(-b).
    \end{equation}
    The three technical conditions are:\footnote{There is actually a fourth technical condition. We require that $R(a)$ and $S(-b)$ only depend on the values $a(x,y)$ and $b(x,y)$ for pairs $(x,y)$ that actually occur with positive probability under $\D$. Equivalently, if $\Pr[a_1(\rn{x}, \rn{y}) = a_2(\rn{x}, \rn{y})] = 1$, then $R(a_1) = R(a_2)$, and a similar condition should hold for $S$.}
    \begin{enumerate}[(a)]
        \item $g$ and $h$ are two-input convex-concave conjugates as in \cref{def:conj-2}.
        \item $R$ and $S$ are one-input convex-concave conjugates as in \cref{def:conj} under
        \[
            \langle a, b \rangle_\D = \E_\D[\langle a(\rn{x}, \rn{y}), b(\rn{x}, \rn{y})\rangle].
        \]
        \item At least one of the functions $g$, $h$, $R$, or $S$ is finite everywhere (i.e. never outputs $\pm \infty$).
    \end{enumerate}
    Also, if the finite function among the four was one of $g,R$, then the supremum on the right-hand side of equation \eqref{thm:six-strong-duality} is attained if finite. Similarly, if the finite function among the four was one of $h,S$, then the infimum on the left-hand side of equation \eqref{thm:six-strong-duality} is attained if finite.
\end{theorem}

\begin{proof}
    In order to apply Fenchel duality (\cref{thm:fenchel}), we must define suitable convex-concave conjugate pairs $g_1,h_1$ and $g_2,h_2$ on the space $\{\X \times \Y \to \R^k\}$.\footnote{Technically, modulo functions that output $0$ with probability $1$, as explained in the previous footnote.} We henceforth refer to this vector space as $\V$, which is finite-dimensional. Indeed, its dimension is precisely $\dim\V = kn$, where the integer $n \le \abs{\X}\abs{\Y}$ is the number of pairs $(x,y) \in \X \times \Y$ that occur with strictly positive probability under $\D$. With this in mind, define
    \begin{align*}
        g_1(a) &= \E[g(\rn{y}, a(\rn{x}, \rn{y}))],\\
        h_1(b) &= \E[h(\rn{y}, b(\rn{x}, \rn{y}))].
    \end{align*}
    It is clear that these are closed proper convex and closed proper concave functions on $\V$, respectively. It is also easy to check that these are conjugates, as per \cref{def:conj}, because for $a,b \in \V$,
    \begin{align*}
        \inf_{a \in \V}\, g_1(a) + \langle a, b \rangle_\D &= \inf_{a \in \V}\, \E[g(\rn{y}, a(\rn{x}, \rn{y})) + \langle a(\rn{x}, \rn{y}), b(\rn{x}, \rn{y})\rangle] \\
        &= \E[\inf_{t \in \R^k}\,g(\rn{y}, t) + \langle t, b(\rn{x}, \rn{y})\rangle] \\
        &= \E[h(\rn{y}, b(\rn{x}, \rn{y}))].
    \end{align*}
    Next, we define $g_2 = R$ and $h_2 = S$. By assumption, these are also a convex-concave conjugate pair of $\eR$-valued functions on $\V$. Finally, we conclude the proof by applying \cref{thm:fenchel} to the pairs $g_1, h_1$ and $g_2, h_2$ defined above.
\end{proof}

\section{Hardness of Learning and Masking Channel Fidelity}
\label{sec:channels}

In this section, we prove \cref{thm:channels}, restated below for convenience, which is our precise characterization of average-case hardness in terms of the fidelity of masking channels. Our proof strategy will be to carefully instantiate our general framework (\cref{thm:six}) with an appropriate choice of the functions $R$ and $S$. The proof relies on various helper lemmas that calculate convex-concave conjugate pairs; we generally defer such lemmas to \cref{sec:calc}.

\getkeytheorem{thm:channels}

\begin{proof}
    Given the functions $g,h$ from the theorem statement, define
    \[
        g_1(y, a) = g(y, \pi(a))
    \]
    and
    \[
        h_1(y, b) = \begin{cases}
            h(y, z) &\text{if $b = z - \vec{1}/k$ for some $z \in \Delta_k$,} \\
            -\infty &\text{otherwise.}
        \end{cases}
    \]
    These are convex-concave conjugates by \cref{thm:uniform-shift-calc}. Next, let $R,S$ be the indistinguishability constraint functionals (\cref{def:hardcore-constraint}), which are convex-concave conjugates by \cref{thm:hardcore-constraint-calc}. Since $h$ has a bounded effective domain, its conjugate $g$, which is defined as a supremum over this domain, is always finite. Hence, so is $g_1$. Therefore, applying \cref{thm:six} to $g_1, h_1, R, S$ yields
    \[
        \inf_{\lambda \in \R^\C}\, \E[g(\rn{y}, \pi(c_\lambda(\rn{x})))] + \eps \|\lambda\|_1 =
        \begin{aligned}[t]
            \max_{z : \X \times \Y \to \Delta_k}\, &\E[h(\rn{y}, z(\rn{x}, \rn{y}))] \\
            \text{s.t.} \; &\forall c \in \C,\; \bigabs{\E[\langle c(\rn{x}), \, z(\rn{x}, \rn{y}) - \vec{1}/k\rangle]} \le \eps.
        \end{aligned}
    \]
    Next, we make a simple notational change. Specifically, we define the noisy channel $\ch : \X \to \CSM$ such that the $j$\textsuperscript{th} column of $\ch(x)$ is $z(x, e_j)$. Then, the optimization problem on the right-hand side of the above equation can be rewritten as
    \[
        \max_{\ch : \X \to \CSM}\, \E[h(\rn{y}, \ch(\rn{x})\rn{y})] \quad\text{s.t.} \quad \forall c \in \C,\; \bigabs{\E[\langle c(\rn{x}), \, \ch(\rn{x})\rn{y} - \vec{1}/k\rangle]} \le \eps.
    \]
    Of course, the constraint on $\ch$ is precisely the requirement that $\ch(\rn{x})\rn{y}$ is $(\C, \eps)$-pseudorandom, as per \cref{def:hardcore-channel}. This proves the main part of \cref{thm:channels}.
    
    For the second part, suppose $g$ is bounded below by the value $v \in \R$ on centered actions. If $\eps > 0$, then it is never beneficial to use  $\lambda$ whose $\ell_1$ norm exceeds
    \[
        r = \frac{\E[g(\rn{y}, 0)] - v}{\eps}.
    \]
    Indeed, any such $\lambda$ is strictly outperformed by $\lambda = 0$. Thus, we can restrict the search space to the $\ell_1$ ball of radius $r$, which is compact since $\C$ is finite. Since the left-hand side minimizes a lower semicontinuous function over a compact set, it attains this minimum.
\end{proof}

Next, we prove a slightly different version of the preceding theorem which uses a \emph{soft} hardcore penalty instead of the hardcore constraint. This version is appealing for its more flexible treatment of the class $\C$ and fewer hyperparameters.

\begin{theorem}[Hardness vs Fidelity, Soft Penalty]
\label{thm:channels-soft}
    For all finite $\X$ and $\Y = [k]$, distributions over $(\rn{x}, \rn{y}) \in \X \times \Y$, compact convex $\overline{\C} \subseteq \{\X \to \R^k\}$, and convex-concave conjugates $g : \Y \times \R^k \to \R$ and $h : \Y \times \Delta_k \to \eR$,
    \[
        \min_{c \in \overline{\C}}\, \E[g(\rn{y}, \pi(c(\rn{x})))] = \max_{\ch : \X \to \CSM}\, \Bigl(\E[h(\rn{y}, \ch(\rn{x})\rn{y})] - \max_{c \in \overline{\C}}\,\E[\langle c(\rn{x}), \, \ch(\rn{x})\rn{y} - \vec{1}/k\rangle]\Bigr).
    \]
\end{theorem}

\begin{proof}
    Define $g_1, h_1$ in terms of $g,h$ as in the preceding proof of \cref{thm:channels}. Let $R,S$ be the hardcore penalty functionals (\cref{def:hardcore-penalty}), which are convex-concave conjugates by \cref{thm:hardcore-penalty-calc}.
    Finally, we note that the function $g_1$ is always finite by assumption, which lets us apply our general duality framework (\cref{thm:six}) to $g_1, h_1, R, S$. This concludes the proof.
\end{proof}

\section{Multiclass Hardcore Lemmas}
\label{sec:applications}

In this section, we state and prove our various instantiations of \cref{thm:channels} and investigate multiclass hardcore certificates via surrogate losses for $\ell_{\text{0-1}}$, as discussed in \cref{sec:intro-applications}. We begin with our result in the binary setting, whose statement we recall for convenience.

\getkeytheorem{thm:binary-hardcores}

\begin{proof}
    Given \cref{thm:channels,thm:six}, the current proof is mostly a matter of defining appropriate loss-fidelity pairs $g, h$ and verifying that they are indeed convex-concave conjugates:
    \begin{itemize}
        \item First, consider the fidelity function
        \[
            h(y, z) = \begin{cases}
                \mu &\text{if $z = \mu y$ for some $\mu \in [0, 1]$,}\\
                -\infty &\text{otherwise.}
            \end{cases}
        \]
        By \cref{def:conj}, its conjugate is
        \[
            \sup_{b \in \R}\, h(y, b) - ab =\sup_{\mu \in [0, 1]}\, \mu - a \cdot (\mu y) = \max(0, 1 - ay),
        \]
        where we have used the fact that an affine function of $\mu \in [0, 1]$ is maximized at either $\mu = 0$ or $\mu = 1$. This is precisely hinge loss.

        \item Next, consider the fidelity function
        \[
            h(y, z) = \begin{cases}
                \mu &\text{if $z = \mu y$ for some $\mu \in [-1, +1]$,}\\
                -\infty &\text{otherwise.}
            \end{cases}
        \]
        By \cref{def:conj}, its conjugate is
        \[
            \sup_{b \in \R}\, h(y, b) - ab =\sup_{\mu \in [-1, +1]}\, \mu - a \cdot (\mu y) = |1 - ay|.
        \]
        Since $y \in \{\pm 1\}$, this is precisely $\ell_1$ loss.
        \item Finally, consider the fidelity function
        \[
            h(y, z) = \begin{cases}
                \mu - \frac{1}{q}\abs{\mu}^q &\text{if $z = \mu y$ for some $\mu \in \R$.}
            \end{cases}
        \]
        We show in \cref{thm:lp-calc} that the conjugate of $h$ is precisely $\ell_p$ loss.
    \end{itemize}
    To conclude the proof, we invoke \cref{thm:channels} with $k=2$. (Technically, we must map distributions in $\Delta_2$, which is a line segment, to the interval $[-1, +1]$, but we omit the details of this very simple change of variables, which is essentially just a change of notation. Also, for the third example, where $\mu$ is not bounded in the interval $[-1, +1]$, we instead apply \cref{thm:six} with $k=2$ and the indistinguishability constraint functionals from \cref{def:hardcore-constraint}.)
\end{proof}

Next, we prove our results regarding the multiclass hardness certificates associated with multiclass hinge losses $\ell_{\mathrm{weak}}$, $\ell_{\mathrm{strong}}$, and $\ell_{\mathrm{CS}}$. We recall the definitions of these loss functions and the statements of the relevant theorems below. For the meaning of these losses and the interpretations of their corresponding hardness certificates, see \cref{sec:intro-applications}.

\begin{definition}[Multiclass Hinge Losses]
Given $a \in \R^k$ with mean $0$, let
\begin{align*}
    \ell_{\mathrm{weak}}(e_i, a) &= (1 - a_i)_+, \\
    \ell_{\mathrm{strong}}(e_i, a) &= \max_{j \in [k]}\,(e_i - a)_j, \\
    \ell_{\mathrm{CS}}(e_i, a) &= (1 - a_i + \max_{j \neq i} a_j)_+.
\end{align*}
We extend these formulas to all $a \in \R^k$ by first centering $a$ to have mean $0$. The third of these hinge-like loss functions, the \emph{Crammer-Singer} loss, was introduced by \cite{crammer2001multiclass}.
\end{definition}
\getkeytheorem{thm:weak-hinge-hardcore}

\begin{proof}
    We apply \cref{thm:channels} with the fidelity function from \cref{ex:fid-mix}, which constrains the masking channel to mix the labels with uniform noise:
    \[
        h(y, z) = \begin{cases}
            \mu &\text{if $z = \mu\cdot y + (1 - \mu) \cdot \vec{1}/k$ for some $\mu \in [0, 1]$,}\\
            -\infty &\text{otherwise.}
        \end{cases}
    \]
    Its conjugate is the loss function
    \begin{align*}
        \sup_z\, h(y, z) - \langle a, z \rangle &= \sup_{\mu \in [0, 1]}\, \mu - \langle a, \mu\cdot y + (1 - \mu)\cdot \vec{1}/k\rangle \\
        &= (1 - \langle a,  y - \vec{1}/k\rangle)_+ - \langle a, \vec{1}/k\rangle,
    \end{align*}
    which agrees with $\ell_{\mathrm{weak}}(y, a)$ for all mean-$0$ inputs $a \in \R^k$.
\end{proof}

\getkeytheorem{thm:strong-hinge-hardcore}

\begin{proof}
We apply \cref{thm:channels} with the fidelity function from \cref{ex:fid-flip}:
\[
    h(y, z) = \begin{cases}
        \langle y, z \rangle &\text{if $z \in \Delta_k$,}\\
        -\infty &\text{otherwise.}
    \end{cases}
\]
Its conjugate is the loss function
\begin{align*}
    \sup_z\, h(y, z) - \langle a, z \rangle &= \sup_{z \in \Delta_k}\, \langle y, z \rangle - \langle a, z\rangle \\
    &= \max_{j \in [k]}\,(y - a)_j,
\end{align*}
which agrees with $\ell_{\mathrm{strong}}(y, a)$ for mean-$0$ actions $a$.
\end{proof}

\getkeytheorem{thm:cs-hinge-hardcore}

\begin{proof}
We apply \cref{thm:six} with the indistinguishability constraints in \cref{def:hardcore-constraint}, which are conjugates by \cref{thm:hardcore-constraint-calc}, as well as the function
\[
    h(y, b) = \begin{cases}
        1 - \langle y, z \rangle &\text{if $b = y - z $ for some $z \in \Delta_k$,}\\
        -\infty &\text{otherwise.}
    \end{cases}
\]
Its conjugate $g(y, a)$ is the function
\begin{align*}
    \sup_b\, h(y, b) - \langle a, b \rangle &= \sup_{z \in \Delta_k}\, 1 - \langle y, z \rangle - \langle a, y - z\rangle \\
    &= 1 - \langle a, y\rangle - \min_{j \in [k]}\,(y - a)_j  \\
    &= (1 - a_y + \max_{j \neq y} a_j)_+,
\end{align*}
which is precisely $\ell_{\mathrm{CS}}(y, a)$.
\end{proof}

\subsection{Bounds for Surrogate Losses}

We now shift gears to our study of surrogate losses for $\ell_{\text{0-1}}$, with the goal of understanding which hardness certificates are guaranteed to exist under the mild assumption of $\delta$-hardness under 0-1 loss for a given class $\C$ after appropriate rounding to $\Delta_k$. As a warm-up before proving our main result showing that surrogate losses can be determined from a loss function's associated entropy, we prove a result specific to the Crammer-Singer loss $\ell_{\mathrm{CS}}$. The result states that 0-1 loss is at most $(k-1)/k$ times the Crammer-Singer hinge loss, provided that we first project $a \in \R^k$ onto the probability simplex $\Delta_k$. While similar bounds appear in the literature \cite{VanDerHoevenFuscoCesaBianchi2021}, we provide a proof for completeness.

\begin{lemma}[
    name = Crammer-Singer Rounding,
    label = thm:cs-round,
    store = thm:cs-round
]
    Given $y \in [k]$ and $a \in \R^k$, let $\tilde{a} = \mathrm{proj}_{\Delta_k}(a)$. Equivalently, let $\tilde{a}_i = (a_i - \tau)_+$ for the unique threshold $\tau \in \R$ such that $\sum_i \tilde{a}_i = 1$. Then,
    \[
        \ell_{\mathrm{CS}}(y, a) \ge \frac{k}{k-1}\E_{\tilde{\rn{y}} \sim \tilde{a}}[\ell_{\textnormal{0-1}}(y, \tilde{\rn{y}})].
    \]
\end{lemma}
\begin{proof}
    Let $y = e_i$. When $\tilde{a}_i = 1$, there's nothing to prove. Otherwise, we have $1 - \tilde{a}_i > 0$, so there must be some $j^\star \neq i$ with $\tilde{a}_{j^\star} \ge (1 - \tilde{a}_i)/(k-1) > 0$. Since $\tilde{a}_{j^\star} = a_{j^\star} - \tau$ and $\tilde{a}_i \ge a_i - \tau$,
    \[
        \ell_{\mathrm{CS}}(e_i, a) \ge 1 - a_i + a_{j^\star} \ge 1 - \tilde{a}_i + \tilde{a}_{j^\star} \ge \frac{k}{k-1}(1 - \tilde{a}_i) = \frac{k}{k-1}\E_{\tilde{\rn{y}} \sim \tilde{a}}[\ell_{\textnormal{0-1}}(e_i, \tilde{\rn{y}})]. \qedhere
    \]
\end{proof}

The preceding result implies, in particular, that $\ell_{\mathrm{CS}}$ is a $k/(k-1)$-surrogate loss for $\ell_{\text{0-1}}$, as formulated in \cref{def:surrogate}. Next, we prove \cref{thm:opt-round}, which shows that $\beta$-surrogate losses can be completely determined by inspecting the relevant losses' associated \emph{entropy} functions, which are given by the formula in \cref{def:entropy}. 

\getkeytheorem{thm:opt-round}

\begin{proof}
    Suppose first that $\ell_\A$ is a $\beta$-surrogate loss for $\ell_\B$ with rounding rule $R$. Then,
    \[
        h_\A(p) = \inf_{a \in \A}\, \E_{\rn{y} \sim p}[\ell_\A(\rn{y}, a)] \ge \beta \inf_{a \in \A}\, \E_{\substack{\rn{y} \sim p \\ \rn{b} \sim R(a)}}[\ell_\B(\rn{y}, \rn{b})] \ge \beta \inf_{b \in \B}\, \E_{\rn{y} \sim p}[\ell_\B(\rn{y}, b)] = \beta h_\B(p).
    \]
    Conversely, suppose that $h_\A(p) \ge \beta h_\B(p)$. Then, by the minimax theorem, for all $a \in \A$,
    \begin{align*}
        \max_{R(a) \in \Delta(\B)} \min_{y \in \Y}\, \E_{\rn{b} \sim R(a)}[\ell_\A(y, a) - \beta \cdot \ell_\B(y, \rn{b})] &= \min_{p \in \Delta(\Y)}\max_{b \in \B} \, \E_{\rn{y} \sim p}[\ell_\A(\rn{y}, a) - \beta \cdot \ell_\B(\rn{y}, b)] \\
        &= \min_{p \in \Delta(\Y)} \, \E_{\rn{y} \sim p}[\ell_\A(\rn{y}, a)] - \beta \cdot h_\B(p) \\
        &\ge \min_{p \in \Delta(\Y)}\, h_\A(p) - \beta \cdot h_\B(p) \\
        &\ge 0.
    \end{align*}
    Note that switching the ``min'' and ``max'' is certainly valid because $\Y$ and $\B$ are finite. Therefore, any maximizer $R(a) \in \Delta(\B)$ for the left-hand side is a rounding rule proving $\beta$-surrogacy.
\end{proof}

In \cref{sec:calc}, we give an alternate proof of \cref{thm:opt-round} in the special case of surrogates for the 0-1 loss. This proof is quite a bit more intuitive than the previous one, and it also clarifies the structure of the optimal rounding rule. Next, we calculate the entropy functions for $\ell_{\text{0-1}}$, $\ell_{\mathrm{weak}}$, $\ell_{\mathrm{strong}}$, and $\ell_{\mathrm{CS}}$ and plug these into \cref{thm:opt-round} to relate each of the losses $\ell_{\mathrm{weak}}$, $\ell_{\mathrm{strong}}$, and $\ell_{\mathrm{CS}}$ to 0-1 loss. This implies, in particular, that $\delta$-hardness under $\ell_{\text{0-1}}$ implies the existence of the hardness certificates corresponding to $\ell_{\mathrm{strong}}$ and $\ell_{\mathrm{CS}}$ with the parameters discussed in \cref{sec:intro-applications}.

\begin{lemma}[Entropy Calculations]
\label{thm:entropy-calc}
    For any $k \ge 2$, we have 
    \begin{align*}
        h_{\textnormal{0-1}}(p) &= 1 - \max_j p_j,\\
        h_{\mathrm{weak}}(p) &= k \min_j p_j,\\
        h_{\mathrm{strong}}(p) &= 1 - d_{\mathrm{TV}}(p, \vec{1/k}),\\ 
        h_{\mathrm{CS}}(p) &= \min\{1, 2(1 - \max_jp_j)\}.
    \end{align*}
\end{lemma}

\begin{corollary}
    If $k \ge 3$, the loss functions $\ell_{\mathrm{weak}}$, $\ell_{\mathrm{strong}}$, and $\ell_{\mathrm{CS}}$ are surrogate losses for $\ell_{\textnormal{0-1}}$ with parameters $\beta_{\mathrm{weak}} = 0$, $\beta_{\mathrm{strong}} = 4/k$, $\beta_{\mathrm{CS}} = k/(k-1)$. If $k = 2$, then all three $\beta$ parameters are equal to $2$ instead. Each of these $\beta$ parameters is optimal.
\end{corollary}

\begin{proof}
    We combine \cref{thm:opt-round} with the formulas from \cref{thm:entropy-calc}.
    \begin{itemize}
        \item For weak hinge loss, take any $p \in \Delta_k$ with some coordinate equal to $0$ and no coordinate equal to $1$ (needs $k \ge 3$). By \cref{thm:entropy-calc}, we have $h_{\mathrm{weak}} = 0$ but $h_{\text{0-1}} > 0$, so $\beta_{\mathrm{weak}} = 0$.

        \item For strong hinge loss, let $m = \max_j p_j$ and note that $p_i \le m$ and $m \ge 1/k$. Therefore,
        \[
            \frac{h_{\mathrm{strong}}(p)}{h_{\text{0-1}}(p)} = \frac{\sum_i \min\{p_i, 1/k\}}{1 - m} \ge \frac{\sum_i p_i/km}{1-m} = \frac{1}{k m(1-m)} \ge \frac{4}{k},
        \]
        with equality at $p = (1/2, 1/2, 0, \ldots, 0)$.
        \item For Crammer-Singer hinge loss, the ratio between $h_{\mathrm{CS}}(p)$ and $h_{\text{0-1}}(p)$ is precisely $\min\{1/(1 - \max_j p_j), 2\}$, which is minimized at the uniform distribution with $\beta_{\mathrm{CS}} = k/(k-1)$.
    \end{itemize}
    The case $k = 2$ is easily checked separately.
\end{proof}

Although we have focused our discussion in the multiclass case on various generalizations of hinge loss that are closely related to 0-1 loss, there are of course many other multiclass convex losses, such as $\ell_p$ losses and more. Each of these gives rise to its own quantitative notion of hardness and, via our results, its own notion of a ``hardness certificate''. We conclude this section with just one such example, pertaining to multiclass $\ell_1$ loss. Here, the hardness certificate is neither a measure nor a masking channel, but a function $\sigma : \X \times \Y \to [-1, +1]^k$, whose outputs need not comprise a probability distribution over $[k]$ (in fact, some coordinates may even be negative).

\begin{theorem}[Multiclass $\ell_1$ Hardcore]
\label{thm:multi-l1}
    For all finite $\X$ and $\Y = [k]$, distributions over $(\rn{x}, \rn{y}) \in \X \times \Y$, finite $\C \subseteq \{\X \to \R^k\}$, and $\eps > 0$,
    \[
        \min_{\lambda \in \R^\C}\, \E[\|\rn{y} - c_\lambda(\rn{x})\|_1] + \eps \|\lambda\|_1 =
        \begin{aligned}[t]
            \max_{\sigma : \X \times \Y \to [-1, +1]^k}\, &\E[\langle \rn{y}, \sigma(\rn{x}, \rn{y})\rangle] \quad \\\text{s.t.} &\; \forall c \in \C,\; \bigabs{\E[\langle c(\rn{x}), \, \sigma(\rn{x}, \rn{y}) \rangle]} \le \eps.
        \end{aligned}
    \]
\end{theorem}

\begin{proof}
    As usual, we apply \cref{thm:six} with the indistinguishability constraints $R,S$ from \cref{def:hardcore-constraint}. We use the functions $g(y, a) = \|y - a\|_1$ and
    \[
        h(y, b) = \begin{cases}
            \langle y, b \rangle &\text{if $b \in [-1, +1]^k$,}\\
            -\infty &\text{otherwise.}
        \end{cases}
    \]
    These are conjugates because
    \begin{align*}
        \sup_{b \in [-1, +1]^k}\, h(y, b) - \langle a, b \rangle = \sup_{b \in [-1, +1]^k}\, \langle y, b \rangle - \langle a, b\rangle = \|y - a\|_1 = g(y, a).
    \end{align*}
    Invoking \cref{thm:six} as usual completes the proof.
\end{proof}

\section{Universal Masking Channels From Multicalibration}
\label{sec:universal}

In this section, we first prove our theorem on universal hardness certificates via masking channels (\cref{thm:universal-informal}). We then show how this result can be instantiated to recover a previous multicalibration-based construction of hardcore measures in the setting of binary 0-1 loss \cite{casacuberta2024complexity,casacuberta2025global}. We state and prove our new result in \cref{sec:universal-multiclass} and compare it to the prior one in \cref{sec:universal-binary}.

\subsection{Deriving the Multiclass Result}
\label{sec:universal-multiclass}

To begin, we need a certain helper lemma. This lemma establishes the key relationship between the \emph{lowest-loss response} $\hat c_g$ to a loss function $g$ and the \emph{highest-fidelity response} $\widehat\ch_h$ to a fidelity function $h$. To begin, we recall the definitions of these post-processing functions:

\getkeytheorem{def:loss-response}
\getkeytheorem{def:fidelity-response}

Having reviewed the definitions of $\hat c_g$ and $\widehat\ch_h$, we now state and prove our key helper lemma, which states that these two functions actually achieve the \emph{same} value for their respective goals:

\begin{lemma}
\label{thm:uniform-redistribution}
    Given a label distribution $z \in \Delta_k$, let $\rn{y}_z \sim z$. Then, for all convex-concave conjugates $g : \Y \times \R^k \to \eR$ and $h : \Y \times \Delta_k \to \eR$, all $\alpha$-optimal $\hat c_g(z)$, and all $\beta$-optimal $\widehat\ch_h(z)$,
    \[
        0 \le \E[g(\rn{y}_z, \hat c_g(z))] - \E[h(\rn{y}_z, \widehat\ch_h(z)\rn{y}_z)] \le \alpha + \beta.
    \]
\end{lemma}

\begin{proof}
    First, define
    \[
        p^\star = \inf_{c(z) \in \R^k}\, \E[g(\rn{y}_z, c(z))] \quad\text{s.t.}\quad \langle c(z), \vec{1} \rangle = 0.
    \]
    Next, define
    \[
        d^\star = \max_{W(z) \in \CSM}\, \E[h(\rn{y}_z, W(z)\rn{y}_z)] \quad \text{s.t.} \quad W(z)z = \frac{\vec{1}}{k}.
    \]
    By \cref{def:loss-response,def:fidelity-response}, the function $\hat c_g$ is $\alpha$-optimal for the optimization problem defining $p^\star$ and $\widehat\ch_h$ is $\beta$-optimal for the optimization problem defining $d^\star$. Therefore, if we could show that $p^\star = d^\star$, we would be done. For this, we just apply \cref{thm:channels} with $\C$ containing just the $k$ functions $c_j(x) = e_j$, so that $c_\lambda(x) = \lambda$ for all $\lambda \in \R^k$, and error parameter $\eps = 0$. We let $c(z) = \pi(\lambda)$. The constraint $W$ is $(\C,0)$-masking for $\rn{y}_z \sim z$ translates to $W(z)z = \vec{1}/k$.
\end{proof}

The preceding lemma corresponds to the important special case of \cref{thm:channels} where the label $\rn{y}_z \in \Y$ is a truly random sample from a distribution $z \in \Delta_k$, entirely independent of the ``context'', ``feature vector'', or ``input'' $x \in \X$. Soon, we will apply this helper lemma to each $z$-level set of a multicalibrated model, within which the true label is \emph{computationally indistinguishable} from precisely this scenario.

We now state and prove the formal version of \cref{thm:universal-informal}, which is the main result of this section. For clarity, we shall split the formal theorem into three parts:
\begin{enumerate}[(1)]
    \item The first part, \cref{thm:universal-properties}, shows that if there exists a model $p : \X \to \Delta_k$ of the conditional distribution $\rn{y}|\rn{x}$ with certain indistinguishability properties depending on the loss function $g$, the fidelity function $h$, and the class $\C$, then $p$ yields a near-optimal noisy channel certifying the hardness of approximating $\rn{y}|\rn{x}$ by members of $\C$.
    
    Implicit in our ``near-optimal'' claim is the assumption that the complexity of $p$ is comparable to that of $\C$. The required complexity bound on $p$ is established in the next part.

    \item The second part, \cref{thm:universal-existence}, shows that even for a large family $\F$ of loss functions $g$ and fidelity functions $h$, there exists a single \emph{universal} model $p : \X \to \Delta_k$ satisfying the relevant indistinguishability properties for all $g,h \in \F$ simultaneously. Moreover, if the functions $c \in \C$ and $g,h \in \F$ and $\hat c_g,\widehat\ch_h$ can all be computed efficiently, then so can $p$.

    The proof of this part will be very simple. It involves a direct appeal to well-known constructions of low-complexity models satisfying \emph{calibration} and \emph{multiaccuracy}, which we reviewed in \cref{sec:prelim-multical}.

    \item The third part, \cref{thm:universal-monotone}, highlights special cases where the noisy channel $\widehat\ch_h(p(x))$ \emph{itself}, not just the model $p(x)$, can be made universal. Specifically, we show that the same post-processing $\widehat\ch = \widehat\ch_h$ works if all fidelity functions $h$ in the class are monotone, or if they are all monotone mixing. For the definition of such loss functions, recall \cref{def:mono-fidelity}.
\end{enumerate}

Before stating the first part of our result, \cref{thm:universal-properties}, we must define the specific indistinguishability conditions that we will require of the model $p$. The first condition is a type of \emph{weighted multiaccuracy} (i.e. a special case of \cref{def:weighted-multiaccuracy}). We shall refer to this condition as \emph{channel-weighted multiaccuracy}:

\begin{definition}[Channel-Weighted Multiaccuracy]
\label{def:channel-multiaccuracy}
    Given finite $\X$ and $\Y = [k]$, a distribution over $(\rn{x}, \rn{y}) \in \X \times \Y$, finite $\C \subseteq \{\X \to \R^k\}$, and a fidelity function response $\widehat\ch_h : \Delta_k \to \CSM$ we say that a model $p : \X \to \Delta_k$ satisfies \emph{$W$-weighted $(\C, \eps)$-multiaccuracy} if for all $c \in \C$,
    \[
        \Bigabs{\E\bigl[\bigl\langle \widehat\ch_h(p(\rn{x}))^\top c(\rn{x}), \, \rn{y} - p(\rn{x})\bigr\rangle\bigr]} \le \eps.
    \]
\end{definition}
Intuition for this condition comes from rewriting the left-hand side as
\[
        \E\bigl[\bigl\langle  c(\rn{x}), \, \widehat\ch_h(p(\rn{x}))\rn{y} - \widehat\ch_h(p(\rn{x})) p(\rn{x})\bigr\rangle\bigr] = \E\bigl[\bigl\langle  c(\rn{x}), \, \widehat\ch_h(p(\rn{x}))\rn{y} - \vec{1}/k\bigr\rangle\bigr].
    \]
    and interpreting this as saying that $W(\x) = \widehat\ch_h(p(\rn{x}))$ is $\C$-masking. 

The second indistinguishability condition that we will require of the model $p$ is a type of \emph{weighted calibration} (i.e. a special case of \cref{def:calibration}). We shall refer to this condition as \emph{duality gap calibration} since the weight functions will be derived from the duality gap in \cref{thm:uniform-redistribution}.

\begin{definition}[Duality Gap Calibration]
\label{def:gap-calibration}
    Given finite $\X$ and $\Y = [k]$, a distribution over $(\rn{x}, \rn{y}) \in \X \times \Y$, finite $\C \subseteq \{\X \to \R^k\}$, a loss function response $\hat c_g : \Delta_k \to \R^k$ and a fidelity function response $\widehat\ch_h : \Delta_k \to \CSM$ we say that a model $p : \X \to \Delta_k$ satisfies \emph{$\eps$-duality gap calibration} if
    \[
        \E\bigl[\bigl\langle \mathrm{Gap}(p(\rn{x})), \, \rn{y} - p(\rn{x})\bigr\rangle\bigr] \le \eps,
    \]
    where we define $\mathrm{Gap}(z) \in \eR^k$ by the formula
    \[
        \mathrm{Gap}(z)_i = g(e_i, \hat c_g(z)) - h(e_i, \widehat\ch_h(z)e_i).
    \]
    For applications, we will typically assume that $\mathrm{Gap}(z)$ is bounded in a box $[-B, B]^k$.
\end{definition}

Intuition for this condition comes from the loss OI technique \cite{gopalan2023lossoi, dwork2021outcome}. Lemma \ref{thm:uniform-redistribution} gives us a guarantee that holds for labels $\rn{y}_p|\rn{x} \sim p(\x)$, the simulated distribution of the predictor $p$. We want this guarantee to hold for the true labels $\y$. Since the tests we wish to {\em fool} only depend on  the predictions, this can be framed as a calibration condition.  

At this point, having defined channel-weighted multiaccuracy and duality gap calibration, we are ready to state and prove the three main results of this section. First, we show that these properties of a model $p : \X \to \Delta_k$ are sufficient for constructing masking channels that certify hardness:

\begin{theorem}[
    label = thm:universal-properties,
    name = Universal Masking Channels -- Properties
]

Fix finite $\X$ and $\Y = [k]$, a distribution over $(\rn{x}, \rn{y}) \in \X \times \Y$, finite $\C \subseteq \{\X \to \R^k\}$, a model $p : \X \to \Delta_k$, and convex-concave conjugates $g : \Y \times \R^k \to \R$ and $h : \Y \times \Delta_k \to \eR$. Let $\hat c_{g} : \Delta_k \to \R^k$ be an $\eps_1$-optimal lowest-loss response, and let $\widehat\ch_h : \Delta_k \to \CSM$ be an $\eps_2$-optimal highest-fidelity response. Then,
    \begin{enumerate}[(a)]
        \item \label{item:universal-feasibility} \textbf{(Hardness $\ge$ Fidelity)} If the model $p$ is $\widehat\ch_h$-weighted $(\C, \eps_3)$-multiaccurate, then the masking channel $\ch_p = \widehat\ch_h \circ p$ derived from $p$ certifies hardness, in the sense that for all $c \in \C$,
        \[
            \E[g(\rn{y}, \pi(c(\rn{x})))] \ge \E[h(\rn{y}, W_p(\rn{x})\rn{y})] - \eps_3.
        \]
        \item \label{item:universal-optimality} \textbf{(Hardness $\le$ Fidelity)} If the model $p$ is $\eps_4$-duality gap calibrated, then the bound from part~\ref{item:universal-feasibility} is nearly tight, in the sense that the action $c_p = \hat c_g \circ p$ achieves loss
        \[
            \E[g(\rn{y}, c_p(\rn{x}))] \le \E[h(\rn{y},  W_p(\rn{x})\rn{y})] + (\eps_1 + \eps_2 + \eps_4).
        \]
        (Note: although $c_p \notin \C$ in general, we show in \cref{thm:universal-existence} that $c_p$ ``nearly'' belongs to $\C$.)
    \end{enumerate}
\end{theorem}
\begin{proof} We prove each part separately.
\begin{enumerate}[(a)]
    \item Convex-concave conjugacy of $g,h$ (\cref{def:conj-2}) says that for all $a,b \in \R^k$,
    \[
        g(y, a) \ge h(y, b) - \langle a, b \rangle.
    \]
    For each $(x,y) \in \X \times \Y$ and $c \in \C$, we can set $a = \pi(c(x))$ and $b = \widehat\ch_h(p(x))y$, which gives
    \[
        g(y, \pi(c(x))) \ge h\bigl(y, \widehat\ch_h(p(x))y\bigr) - \bigl\langle \pi(c(x)), \, \widehat\ch_h(p(x))y \bigr\rangle.
    \]
    Our goal is to show that the inner product term on the right is small in expectation. To begin, let us rewrite it slightly. Since the centering function $\pi(\cdot)$ is equivalent to subtracting off the mean of $c(x)$, we can rewrite this inner product as
    \[
        \bigl\langle \pi(c(x)), \, \widehat\ch_h(p(x))y \bigr\rangle = \bigl\langle c(x), \, \widehat\ch_h(p(x))y - \vec{1}/k \bigr\rangle.
    \]
    Next, we make the substitution $\vec{1}/k = \widehat\ch_h(z)z$, where we define $z = p(x)$. This is valid by the very definition of $\widehat\ch_h$ (\cref{def:fidelity-response}). Making this substitution yields the expression
    \[
        \bigl\langle c(x), \, \widehat\ch_h(p(x))(y - p(x)) \bigr\rangle.
    \]
    Next, we move the factor of $\widehat\ch_h(p(x))$ to the other side of the inner product:
    \[
        \bigl\langle \widehat\ch_h(p(x))^\top c(x), \, y - p(x) \bigr\rangle.
    \]
    By the $\widehat\ch_h$-weighted $(\C, \eps_3)$-multiaccuracy assumption, in expectation, this quantity is
    \[
        \E\bigl[\bigl\langle \widehat\ch_h(p(\rn{x}))^\top c(\rn{x}), \, \rn{y} - p(\rn{x}) \bigr\rangle\bigr] \le \eps_3.
    \]

    \item Our goal is to show that
    \[
        \E[g(\rn{y}, \hat c_g(p(\rn{x})))] - \E[h(\rn{y}, \widehat\ch_h(p(\rn{x}))\rn{y})] \le \eps_1 + \eps_2 + \eps_4.
    \]
    Alternatively, phrased in terms of the function $\mathrm{Gap}$ from \cref{def:gap-calibration}, we want
    \[
        \E[\langle \mathrm{Gap}(p(\rn{x})), \rn{y} \rangle] \le \eps_1 + \eps_2 + \eps_4.
    \]
    By the $\eps_4$-duality gap calibration assumption, it suffices to show that
    \[
        \E[\langle \mathrm{Gap}(p(\rn{x})), p(\rn{x}) \rangle] \le \eps_1 + \eps_2.
    \]
    But we have
    \begin{align*}
        \E[\langle \mathrm{Gap}(p(\rn{x})), p(\rn{x}) \rangle] &= \E_{\rn{x}}\Bigl[\sum_{i \in [k]}p(\rn{x})_i \mathrm{Gap}(p(\rn{x}))_i\Bigr] \\
        &= \E_{\rn{x}}\Bigl[\sum_{i \in [k]}p(\rn{x})_i(
        g(e_i, \hat c_g(p(\rn{x}))) - h(e_i, \widehat\ch_h(p(\rn{x}))e_i))\Bigr]\\
        &= \E_{\rn{x},\,\rn{y}_p|\rn{x} \sim p(\x)}\bigl[g(\rn{y}_p, \hat c_g(p(\rn{x}))) -h(\rn{y}_p, \widehat\ch_h(p(\rn{x}))\rn{y}_p)\bigr].
    \end{align*}
    So the desired bound follows from \cref{thm:uniform-redistribution}. 
\end{enumerate}
\end{proof}

At this point, we have shown that if the model $p : \X \to \Delta_k$ satisfies weighted calibration and weighted multiaccuracy, it yields our desired universal masking channel. The next theorem asserts that such models indeed exist for a large family of loss functions $g$ and fidelity functions $h$ \emph{simultaneously}. Perhaps the best way to interpret the following statement, which expresses $p$ as a combination of various functions $g,h,\hat c_g, \widehat\ch_h$, is as follows: if the $g,h,\hat c_g, \widehat\ch_h$ are efficiently computable, which they generally should be, then the predicted label distribution $p(x) \in \Delta_k$ is computable by combining the outputs of $O(\log(k)/\eps^2)$ functions from the class $\C$ on the input $x$. In other words, the complexity of $p$ exceeds that of $\C$ by only an $O(\log(k)/\eps^2)$ factor.

\begin{theorem}[
    label = thm:universal-existence,
    name = Universal Masking Channels -- Existence
]
In the setting of \cref{thm:universal-properties}, consider a family of tuples $(g, h, \hat c_g, \widehat\ch_h)$, where $g,h$ are convex-concave loss-fidelity pairs and $\hat c_g, \widehat\ch_h$ are (possibly suboptimal) responses. Let $\F$ be the collection of all functions $f$ with one of two possible forms:
\begin{itemize}
    \item $f(z, x) = \widehat\ch_h(z)^\top c(x)$ for some $c \in \C$ and some $\widehat\ch_h$ in the family (see \cref{def:channel-multiaccuracy}),
    \item $f(z, x) = \mathrm{Gap}(z)$ for some $g,h,\hat c_g,\widehat\ch_h$ in the family (see \cref{def:gap-calibration}).
\end{itemize}
Suppose that the range of each $f \in \F$ is contained in the box $[-B, +B]^k$. Then, there exists a single model $p : \X \to \Delta_k$ satisfying all of the assumptions in \cref{thm:universal-properties} for all loss-fidelity pairs $g,h$ in the family. Moreover, $p$ has the form $p = p_m$, where we define $p_0(x) = \vec{1}/k$ and
\[
    p_j(x) \propto \exp\Bigl(\frac{\eps}{B^2} \sum_{i=1}^{j} f_i(p_{i-1}(x), x)\Bigr),
\]
normalized so that $\sum_y p_j(x)_y = 1$, for some $m = O(B^2\log(k)/\eps^2)$ functions $f_1, \ldots, f_m \in \pm\F$, and for some common error parameter $\eps = \Omega(\min\{\eps_1, \ldots, \eps_4\})$, where $\eps_i$ is defined in \cref{thm:universal-properties}.
\end{theorem}
\begin{proof}
    Apply \cref{thm:multicalibration} to $\F$ scaled down by a factor of $B$, as well as $\eps/B$.
\end{proof}

\begin{theorem}[
    label = thm:universal-monotone,
    name = Universal Masking Channels -- Common Post-Processing
]
     There exists a common highest-fidelity response function $\widehat\ch = \widehat\ch_h$ for all monotone fidelity functions. There exists a (different) common highest-fidelity response function $\widehat\ch = \widehat\ch_h$ for all monotone mixing fidelity functions.
\end{theorem}

\begin{proof}
    The proof is essentially exactly as we sketched in the introduction and depicted in Figure~\ref{fig:probability-redistribution}. Indeed, let $\widehat\ch_1(z) \in \CSM$ be any noisy channel satisfying the following properties:
\begin{itemize}
    \item \textbf{(Feasible)} $\widehat\ch_1(z)z = \vec{1}/k$. That is, $\widehat\ch_1$ adjusts the probability mass of $z$ to be uniform.
    \item \textbf{(Minimal Movement)} For all $i,j \in [k]$, write the probability mass moved from $i$ to $j$ as
    \[
        m_{ij} = z_i \cdot \langle \widehat\ch_1(z) e_i, e_j \rangle.
    \]
    For all $i$, we require $\sum_{j \neq i} m_{ij} = (z_i - 1/k)_+$, and for all $j$, we require $\sum_{i \neq j} m_{ij} = (1/k - z_j)_+$.
\end{itemize}
It is clear that such a function $\widehat\ch_1$, visualized in Figure~\ref{fig:excess-redistribution}, maximizes $\E[h(\rn{y}, \widehat\ch_1(z)\rn{y})]$ over $\rn{y} \sim z$ subject to the feasibility constraint for any monotone fidelity function $h$, as defined in \cref{def:mono-fidelity}. Similarly, for monotone mixing fidelity functions, also defined in \cref{def:mono-fidelity}, let $z_{\mathrm{min}} = \min_j z_j$ and consider the matrix $\widehat\ch_2(z)$ whose $i$\textsuperscript{th} column is
\[
    \frac{z_{\mathrm{min}}}{z_i}e_i + \Bigl(1 - \frac{z_{\mathrm{min}}}{z_i}\Bigr)\frac{\vec{1}}{k}.
\]
Intuitively, this channel takes all probability mass initially on label $i$ above the minimum and redistributes it uniformly among the $k$ labels. Clearly, $\widehat\ch_2$ is feasible and maximizes $\E[h(\rn{y}, \widehat\ch_2(z)\rn{y})]$ for any monotone mixing fidelity function $h$. This concludes the proof.
\end{proof}

\subsection{Recovering the Binary Result}
\label{sec:universal-binary}

In this section, we show how \cref{thm:universal-properties,thm:universal-existence,thm:universal-monotone} can be used to recover a recent result of \cite{casacuberta2024complexity,casacuberta2025global}, which shows how to construct a hardcore measure in the traditional setting of binary 0-1 loss given a multicalibrated (or calibrated multiaccurate) predictor $p$. In fact, our new \cref{thm:universal-monotone} shows that the construction of this recent result actually works for \emph{all} monotone fidelity functions in the binary setting.

\begin{theorem}[\cite{casacuberta2024complexity,casacuberta2025global}]
\label{thm:cgkr-construction}
For all finite $\X$, distributions over $(\rn{x}, \rn{y}) \in \X \times \{\pm1\}$, finite $\C \subseteq \{\X \to \R\}$, and models $p : \X \to [-1, +1]$, define $\mu : \{\pm 1\} \times [-1, +1] \to [0, 1]$ by
\[
    \mu(y, z) = \frac{1 - yz}{1 + \abs{z}}.
\]
Then, the following statements hold:
\begin{enumerate}[(a)]
        \item \textbf{(Hardness $\ge$ Density)} If $p$ satisfies the weighted multiaccuracy condition
        \[
            \forall c \in \C,\qquad \E\Bigl[\frac{c(\rn{x})}{1 + \abs{p(\rn{x})}}\bigl(\rn{y} - p(\rn{x})\bigr)\Bigr] \le \eps_2,
        \]
        then the measure $\mu$ derived from $p$ certifies hardness against $\C$:
        \[
            \forall c \in \C,\qquad \E[(1 - c(\rn{x})\rn{y})_+] \ge \E[\mu(\rn{y}, p(\rn{x}))] - \eps_2.
        \]
        \item \textbf{(Hardness $\le$ Density)}  If $p$ satisfies the calibration condition
        \[
            \E\Bigl[\frac{\sign(p(\rn{x}))}{1 + \abs{p(\rn{x})}}\bigl(\rn{y} - p(\rn{x})\bigr)\Bigr] \ge - \eps_3,
        \]
        then the bound from part~\ref{item:universal-feasibility} is nearly tight:
        \[
            \E[(1 - \sign(p(\rn{x}))\rn{y})_+] \le \E[\mu(\rn{y}, p(\rn{x}))] + \eps_3.
        \]
    \end{enumerate}
\end{theorem}

\begin{proof}
    In the statements of \cref{thm:universal-properties,thm:universal-existence,thm:universal-monotone}, let us substitute $k = 2$ and map the set $\Delta_2$, which is a line segment, to the interval $[-1, +1]$ in the natural way. In particular, the highest-fidelity response $\widehat\ch(z) \in \kCSM{2}$, which was previously a $2$-by-$2$ matrix with columns in $\Delta_2$, is now just a pair of scalars $\mu(y, z) \in [-1, +1]$ for $y \in \{\pm 1\}$. We must verify three properties:
    \begin{enumerate}[(i)]
        \item The highest-fidelity response simplifies to
        \[
            \mu(y, z) = \frac{1 - yz}{1 + \abs{z}}.
        \]
        \item The channel-weighted multiaccuracy distinguishers simplify to
        \[
            \frac{c(x)}{1 + \abs{p(x)}}.
        \]
        \item The duality gap calibration distinguishers simplify to
        \[
            -\frac{\sign(p(x))}{1 + \abs{p(x)}}.
        \]
    \end{enumerate}
    To begin, consider the fidelity function from \cref{ex:fid-mix}, specialized to $k=2$ and $\pm 1$ notation:
    \[
        h(y, z) = \begin{cases}
            \mu &\text{if $z = \mu y$ for some $\mu \in [0, 1]$,}\\
            -\infty &\text{otherwise.}
        \end{cases}
    \]
    Its convex conjugate $g$ is hinge loss, whose lowest-loss response is $\hat c_g(z) = \sign(z)$. 
    We now check the three properties listed earlier:
    
    \begin{enumerate}[(i)]
        \item First, to determine the highest-fidelity response, which we write as $\mu(y, z)$, we could appeal to \cref{thm:universal-monotone} by monotonicity, but for clarity, let us calculate it directly using \cref{def:fidelity-response}. In $\pm 1$ notation, we seek the function $\mu : \{\pm 1\} \times [-1, +1] \to [0, 1]$ that maximizes
        \[
            \E_{\rn{y} \sim z}[\mu(\rn{y}, z)]
        \]
        subject to the constraint
        \[
            \E_{\rn{y} \sim z}[\mu(\rn{y}, z)\rn{y}] = \frac{1 + z}{2}\mu(+1, z) - \frac{1-z}{2}\mu(-1, z) = 0.
        \]
        From the constraint, we immediately see that $\mu$ must have the form
        \[
            \mu(y, z) = (1-yz)\nu(z)
        \]
        for some function $\nu : [-1, +1] \to [0, +\infty)$. In terms of $\nu$, our objective is to maximize
        \[
            \E_{\rn{y} \sim z}[(1 - \rn{y}z)\nu(z)] = (1 - z^2)\nu(z).
        \]
        Thus, we should take $\nu$ to be as large as possible. However, we also need the range of $\mu$ to remain bounded in the interval $[0, 1]$. Thus, we choose $\nu = 1/(1 + \abs{z})$, which leads to
        \[
            \mu(y, z) = \frac{1-yz}{1 + \abs{z}}.
        \]
        \item The channel-weighted multiaccuracy test
        \[
            \bigl\langle \widehat\ch_h(p(x))^\top c(x), \, y - p(x) \bigr\rangle,
        \]
        which can be rewritten as
        \[
            \bigl\langle c(x), \, \widehat\ch_h(p(x)) y - \vec{1}/2\bigr\rangle,
        \]
        simplifies substantially in $\pm 1$ notation to
        \[
            c(x) \cdot \mu(y, p(x)) y = \frac{c(x)}{1 + \abs{p(x)}}(y - p(x)),
        \]
        as desired.
        \item The duality gap calibration test
        \[
            \langle \mathrm{Gap}(p(x)), \, y - p(x) \rangle,
        \]
        which can be rewritten as
        \[
            g(y, \hat c_g(p(x))) - h(y, \mu(y, p(x))y),
        \]
        simplifies substantially in $\pm 1$ notation to
        \[
            (1 - \sign(p(x))y)_+ - \mu(y, p(x)) = -\frac{\sign(p(x))}{1 + \abs{p(x)}}(y - p(x)),
        \]
        as desired.
    \end{enumerate}
\end{proof}

\section{Characterizing Pseudoentropy via MaxEnt}
\label{sec:maxent}

In this section, we show that an exact characterization of pseudoentropy can be obtained using the MaxEnt algorithm. 
As discussed in the introduction, this result brings together two quite distinct lines of work: one in complexity theory \cite{vadhan2012pseudoentropy,zheng2014thesis,hu2025generalized} and another in machine learning \cite{berger1996maxent,lafferty2000bregman,dudik2007maxent,altun2006maxent}. We also show that it
can be derived from our general \cref{thm:six}. 

To begin, let us revisit the starting point of our investigation: the classic hardcore lemma for binary 0-1 loss (\cref{thm:classic-hardcore}). Recall that the first step we took, even before moving to the multiclass case, was to take the $2\delta$-hardcore measure $\mu : \X \to [0,1]$ and use it to decompose the function $f : \X \to \{\pm 1\}$ into
\[
    f(x) = \underbrace{\mu(x) f(x)}_{\text{noisy labels}} + \underbrace{(1 - \mu(x))f(x)}_{\text{high-entropy model}}.
\]
The first term, $\mu(x)f(x)$, which randomizes $f$ outside the hardcore set, was the inspiration for our new \emph{noisy channel} property. However, the other term, $(1 - \mu(x))f(x)$, which randomized $f$ inside the hardcore, is the starting point for the investigation of generalized \emph{pseudoentropy}, pioneered by \cite{vadhan2012pseudoentropy,vadhan2013uniform}. The relation between these two views can be visualized using Figure \ref{fig:maxent-vs-hardcore}.

We now state a result that exactly characterizes pseudoentropy via (the hardness of) minimizing the associated GLM loss. This provides an analogue of our Theorem \ref{thm:channels} in the setting of pseudoentropy. 

\getkeytheorem{thm:maxent}

\begin{proof}
    Given $g,h,\ell$ from the theorem statement, let $g_1 = \ell$ and
    \[
        h_1(y, b) = \begin{cases}
            h(z) &\text{if $b = y - z$ for some $z \in \Delta_k$,} \\
            -\infty &\text{otherwise.}
        \end{cases}
    \]
    By \cref{thm:glm-calc}, the functions $g_1$ and $h_1$ are conjugates.
    Next, let $R,S$ be the indistinguishability constraint functions from \cref{def:hardcore-constraint}, which are also conjugates by \cref{thm:hardcore-constraint-calc}. Since $h$ has a bounded effective domain, its conjugate $g$, which is defined as a supremum over this domain, is always finite. Hence, so is $g_1$. Therefore, applying \cref{thm:six} to $g_1, h_1, R, S$ yields
    \[
        \inf_{\lambda \in \R^\C}\, \E[\ell(\rn{y}, c_\lambda(\rn{x}))] + \eps \|\lambda\|_1 =
        \begin{aligned}[t]
            \max_{p : \X \times \Y \to \Delta_k}\, &\E[h(p(\rn{x}, \rn{y}))] \\
            \text{s.t.} \; &\forall c \in \C,\; \bigabs{\E[\langle c(\rn{x}), \, \rn{y} - p(\rn{x}, \rn{y})\rangle]} \le \eps.
        \end{aligned}
    \]
    
    Next, we show that without loss of generality, we may assume that the function $p(x, y)$ on the right-hand side ignores its second input. To this end, define
    \[
        \bar{p}(x_0) = \E[p(\rn{x}, \rn{y}) \mid \rn{x} = x_0] \in \Delta_k.
    \]
    Clearly, if $p$ is $(\C, \eps)$-multiaccurate (i.e. feasible for the maximization problem) then so is $\bar{p}$. Moreover, by concavity of $h$ and Jensen's inequality, we have
    \[
        \E[h(p(\rn{x}, \rn{y}))] \le \E[h(\bar{p}(\rn{x}))].
    \]
    Thus, replacing $p$ with $\bar{p}$, we see that $y$-dependence in the maximizer is unnecessary.
    
    The proof of the final claim is almost exactly the same as in the proof of \cref{thm:channels}, and we include it for the sake of completeness: Suppose $\ell$ is bounded below by the value $v \in \R$. If $\eps > 0$, then it is never beneficial to use  $\lambda$ whose norm exceeds
    \[
        r = \frac{\E[\ell(\rn{y}, 0)] - v}{\eps}.
    \]
    Indeed, any such $\lambda$ is strictly outperformed by $\lambda = 0$. Thus, we can restrict the search space to the $\ell_1$ ball of radius $r$, which is compact since $\C$ is finite. Since the left-hand side minimizes a lower semicontinuous function over a compact set, it attains this minimum.
\end{proof}

\bibliographystyle{alpha}
\bibliography{sources}
\appendix
\section{Additional Calculations}
\label{sec:calc}

\begin{lemma}
\label{thm:uniform-shift-calc}
    Given convex-concave conjugates $g : \Y \times \R^k \to \R$ and $h : \Y \times \Delta_k \to \R$, define
    \[
        g_1(y, a) = g(y, \pi(a))
    \]
    and
    \[
        h_1(y, b) = \begin{cases}
            h(y, z) &\text{if $b = z - \vec{1}/k$ for some $z \in \Delta_k$,} \\
            -\infty &\text{otherwise.}
        \end{cases}
    \]
    Then $g_1, h_1$ are also convex-concave conjugates.
\end{lemma}
\begin{proof}
    We calculate the conjugate of $h_1$ as follows:
    \begin{align*}
        \sup_{b \in \R^k}\, h_1(y, b) - \langle a, b \rangle
        &=\sup_{z \in \Delta_k}\, h(y, z) - \langle a, z - \vec{1}/k \rangle \\
        &=\sup_{z \in \Delta_k}\, h(y, z) - \langle \pi(a), z \rangle \\
        &= g(y, \pi(a)). \qedhere
    \end{align*}
\end{proof}

\begin{lemma}
\label{thm:hardcore-constraint-calc}
    The indistinguishability constraint functionals (\cref{def:hardcore-constraint}) are conjugates.
\end{lemma}

\begin{proof}
    We first recall the definitions of $R$ and $S$:
    \begin{align*}
        R(a) &= \inf_{\lambda \in \R^\C} \, \eps \|\lambda\|_1  \text{ such that $\Pr\Bigl[a(\rn{x}, \rn{y}) = \sum_i \lambda_i c_i(\rn{x})\Bigr] = 1$} \\
        S(b) &= \begin{cases}
            0 &\text{if $\abs{\E[\langle c(\rn{x}), b(\rn{x},\rn{y})\rangle]} \le \eps$ for all $c \in \C$,} \\
            -\infty &\text{otherwise.}
        \end{cases}
    \end{align*}
    Note that $R(a) = +\infty$ if $a \notin \spn(\C)$.
    To verify that $R$ and $S$ are conjugates, we first write down the expression for the conjugate of $R$, as described in \cref{def:conj}:
    \begin{align*}
        \inf_{\lambda}\, \eps\|\lambda\|_1 + \E[\langle c_\lambda(\rn{x}), b(\rn{x}, \rn{y})\rangle].
    \end{align*}
    This quantity is clearly $-\infty$ if there exists any $c^* \in \C$ with $\abs{\E[\langle c^*(\rn{x}), b(\rn{x}, \rn{y})\rangle]} > \eps$. Indeed, just let the coefficient vector $\lambda$ assign increasingly large positive (or negative) values to the coordinate corresponding to $c^*$, while assigning $0$ to all other coordinates. Otherwise, if $\abs{\E[\langle c(\rn{x}), b(\rn{x}, \rn{y})\rangle]} \le \eps$ for every $c \in \C$, the overall quantity is $0$, which is attained by the optimizer $\lambda = 0$. This is precisely the function $S$ that we just defined.
\end{proof}

\begin{lemma}
\label{thm:hardcore-penalty-calc}
    The indistinguishability penalty functionals (\cref{def:hardcore-penalty}) are conjugates.
\end{lemma}

\begin{proof}
    We first recall the definitions of $R$ and $S$:
    \begin{align*}
        R(a) &= \begin{cases}
            0 &\text{if $\Pr[a(\rn{x}, \rn{y}) = c(\rn{x})] = 1$ for some $c \in \overline{\C}$,}\\
            +\infty &\text{otherwise.}
        \end{cases}\\
        S(-b) &= -\max_{c \in \overline{\C}}\, \E[\langle c(\rn{x}), b(\rn{x},\rn{y})\rangle].
    \end{align*}
    To verify that $R$ and $S$ are conjugates, we simply write down the expression for the conjugate of $R$, as described in \cref{def:conj}. Doing so gives 
    \(
        \min_{c \in \overline{\C}}\, \E[\langle c(\rn{x}),b(\rn{x}, \rn{y})\rangle],
    \)
    which is just $S(b)$ for the function $S$ that we just defined.
\end{proof}

\begin{lemma}
\label{thm:lp-calc}
Let $p,q>1$ satisfy $1/p + 1/q = 1$. Then $\ell_p(y, a) = (1/p)\abs{y-a}^p$ and
\[
    h(y, z) = \begin{cases}
        \mu - \frac{1}{q}\abs{\mu}^q &\text{if $z = \mu y$ for some $\mu \in \R$}
    \end{cases}
\]
are convex-concave conjugates, assuming $y \in \{\pm 1\}$.
\end{lemma}
\begin{proof}
To see that these are conjugates, let us calculate
\begin{align*}
    \sup_{b \in \R}\, h(y, b) - ab =\sup_{\mu \in \R}\, \mu - \frac{1}{q}\abs{\mu}^q- a \cdot (\mu y)
\end{align*}
Taking the derivative of the right-hand side objective, we see that it is maximized at the point $\mu^\star = \sign(1-ay)\abs{1-ay}^{1/(q-1)}$. Completing the calculation with $\mu^\star$, we see that
\begin{align*}
    \sup_{b \in \R}\, h(y, b) - ab &= \mu^\star - \frac{1}{q} \abs{\mu^\star}^q - a \cdot (\mu^\star y),\\
    &= (1 - ay)\mu^\star - \frac{1}{q}\abs{\mu^\star}^q \\
    &= (1-ay) \cdot \sign(1-ay)\abs{1-ay}^{1/(q-1)} - \frac{1}{q}\abs{1-ay}^{q/(q-1)}\\
    &= \Bigl(1 - \frac{1}{q}\Bigr)\abs{1-ay}^{q/(q-1)}\\
    &= \frac{1}{p}\abs{1-ay}^{p}\\
    &= \frac{1}{p}\abs{y-a}^p,
\end{align*}
where, in the last step, we used the fact that $y \in \{\pm 1\}$.
\end{proof}

Next, we give an alternate proof of \cref{thm:opt-round} in the special case of surrogates for the 0-1 loss. This proof is quite a bit more intuitive than the one we gave in \cref{sec:applications}, and it also clarifies the structure of the optimal rounding rule. As in our first proof, the non-trivial direction is to show the existence of a rounding scheme that matches the entropy gap. Before starting the proof, we make some simplifying observations. 

\begin{itemize}

\item We have not assumed that the losses are non-negative, but when $\ell_\B = \ell_{\text{0-1}}$, it is an implicit requirement for either inequality to hold with $\beta > 0$, since both $h_{\text{0-1}}$ and $\ell_{\text{0-1}}$ are non-negative. For instance, if $\ell(y,a) < 0$, then clearly $h(y) \leq \ell(y,a) < 0$, so $\ell$ cannot be a $\beta$-surrogate for $\ell_{\text{0-1}}$ for any $\beta > 0$. This is not a strong restriction since natural losses and entropies are non-negative.

\item Scaling the loss $\ell = \ell_\A$ by a constant positive factor also scales the entropy $h$. So after scaling both by $1/\beta$, we may in fact assume $\beta =1$. 

\item The non-trivial direction is to come up with a rounding scheme that matches the entropy gap. With the rescaling $\beta =1$, the guarantee $\ell(y,a) \geq \E_{\tilde{\rn{y}} \sim R(a)}[\ell_{\text{0-1}}(y, \tilde{\rn{y}})]$ becomes trivial if $\ell(y,a) \geq 1$. This suggests that the rounding $R(a)$ focus on labels where $\ell(i, a) \in [0,1]$. 

\item Since the probability $R(a)_i$ should decrease with loss, this naturally suggests the quantities $(1 - \ell(i, a))_+$. Indeed, the proof will show that they can be used as probabilities. The rest of the mass can be distributed arbitrarily.
\end{itemize}

\begin{proof}[Alternate Proof of \cref{thm:opt-round} when $\ell_\B = \ell_{\textnormal{0-1}}$]
As in our previous proof of \cref{thm:opt-round}, the surrogate loss relation implies the entropy relation by monotonicity. In the other direction, assume $\ell = \ell_\A$ and $\ell_{\text{0-1}}$ satisfy $h(p) \ge \beta \cdot h_{\text{0-1}}(p)$ for $\beta =1$. For each $i \in [k]$, define the quantity
\[ q_i = (1 - \ell(i, a))_+\]
and let \(Q = \sum_i q_i.\)
We first show that $Q \leq 1$. Let $S = \{i: \ell(e_i,a) < 1\}$, and let $m = |S|$. We may assume $S$ is non-empty, else $\ell(i,a) \geq 1$ for all $i$ and the claimed bound holds trivially.  
We consider $p = u_S$, the uniform distribution over $S$. Then,
\[ 1 - \frac{1}{m} = h_{\text{0-1}}(u_S) \leq h(u_S) \leq \E_{\rn{y} \sim u_S}[\ell(\rn{y}, a)] = \frac{1}{m}\sum_{i \in S} \ell(i, a). \]
Hence
\[ \sum_{i \in S}\ell(i,a) \geq m - 1.\]
We use this to bound
\[ Q = \sum_i(1 - \ell(i,a))_+ = \sum_{i \in S}(1 - \ell(i,a)) = m - \sum_{i \in S}\ell(i,a) \leq 1. \]
Our rounding $R(a)$ will distribute $Q$ probability mass according to $q_i$ and the remaining $1 -Q$ mass arbitrarily, so that $R(a)_i \geq q_i$ for every $i \in [k]$. So we can bound the 0-1 loss by
\[ \ell_{\text{0-1}}(i,R(a)) = 1 - R(a)_i \leq 1- q_i \leq \ell(i,a). \]
where the last inequality holds since
\[
    1- q_i = 1 - (1 - \ell(i,a))_+ \leq \ell(i,a). \qedhere
\]
\end{proof}

\begin{proof}[Proof of \Cref{thm:entropy-calc}]\;
\begin{itemize}
    \item For 0-1 loss, it's clear the best action is $\argmax_j p_j$, which incurs loss $1 - \max_j p_j$.
    \item For weak hinge loss, let $j^\star = \argmin_j p_j$ be the least likely label. The action $a = \vec{1} - ke_{j^\star}$ incurs loss $k \min_j p_j$. To see that this is optimal, consider any $a \in \R^k$ with mean $0$. Then,
    \[
        \sum_{i=1}^k p_i (1 - a_i)_+ \ge \min_j p_j \cdot \sum_{i=1}^k (1 - a_i)_+ \ge \min_j p_j \sum_{i=1}^k(1 - a_i) = k \min_jp_j.
    \]
    \item For strong hinge loss,  let $S = \{i \in [k] \mid p_i > 1/k\}$ and consider the action $a = \bm{1}_S\in \{0, 1\}^k$. After subtracting $\abs{S}/k$ from each coordinate, it has mean $0$ and incurs loss
    \[
        \sum_{i=1}^k p_i \max_j\,(e_i - a)_j + \frac{\abs{S}}{k}= \sum_{i\notin S} p_i + \frac{\abs{S}}{k}.
    \]
    This is just $\sum_i \min\{p_i, u_i\}$ where $u$ is the uniform distribution, or equivalently $1 - d_{\mathrm{TV}}(p, u)$. To see that this is tight, for any $a \in \R^k$, define the margins $t_i = a_i - \min_j a_j$. It is clear that $a,t \in \R^k$ incur the same loss. Using the fact that $t_i \ge 0$, this loss is
    \[
        \sum_{i=1}^k p_i\max_j(e_i - t)_j + \frac{1}{k}\sum_{i=1}^k t_i = \sum_{i=1}^k \Bigl(p_i(1 - t_i)_+ + \frac{t_i}{k}\Bigr).
    \]
    Each summand is minimized at either $t_i = 0$ or $t_i = 1$, and picking the better of the two yields $\sum_i \min\{p_i, u_i\}$, as claimed.
    
    \item For Crammer-Singer hinge loss, consider any $a \in \R^k$ and let $j^\star \in \argmax_j a_j$ and let $m = a_{j^\star} - \max_{j \neq j^\star} a_j$ be the margin between the top two coordinates of $a$. If the label is $j^\star$, then $a$ incurs loss $(1 - m)_+$, and otherwise $a$ incurs loss $1 + m$ or more. Therefore,
    \[
        \sum_{i=1}^k p_i \ell_{\mathrm{CS}}(e_i, a) \ge p_{j^\star}(1 - m)_+ + (1 - p_{j^\star})(1 + m).
    \]
    As a function of $m \ge 0$, this is minimized at either $m = 0$ or $m = 1$, which yield loss values $1$ and $2(1 - p_{j^\star})$, respectively, attained by $a = 0$ and $a = e_{j^\star}$ after centering. \qedhere
\end{itemize}
\end{proof}

\begin{lemma}
\label{thm:glm-calc}
    If $g,h : \R^k \to \eR$ are convex-concave conjugates, then so are
    \[
        g_1(y, a) = g(-a) - \langle y, a \rangle
    \]
    and
    \[
        h_1(y, b) = h(y - b).
    \]
\end{lemma}
\begin{proof}
    We calculate the conjugate of $h_1$ as follows:
    \begin{align*}
        \sup_{b \in \R^k}\, h_1(y, b) - \langle a, b \rangle
        &=\sup_{z \in \R^k}\, h(z) - \langle a, y - z \rangle \\
        &=\sup_{z \in \R^k}\, \bigl(h(z) + \langle a, z \rangle\bigr) - \langle y, a \rangle \\
        &= g(-a) - \langle y, a \rangle.
    \end{align*}
    By definition, this is precisely $g_1$.
\end{proof}
\end{document}